\documentclass[letterpaper]{article}
\usepackage[preprint]{aaai2027}
\usepackage[hyphens]{url}
\usepackage{graphicx}
\usepackage{natbib}
\usepackage{amsmath}
\usepackage{amssymb}
\usepackage{caption}
\usepackage[linesnumbered,ruled,vlined]{algorithm2e}
\usepackage{booktabs}
\usepackage{multirow}
\usepackage{threeparttable}
\usepackage{tikz}
\usetikzlibrary{arrows.meta,calc,patterns}
\usepackage{pifont}

\usepackage{xcolor}
\usepackage{listings}
\usepackage[most]{tcolorbox}
\usepackage[varqu,scaled=0.94]{zi4}

\newcommand{\coderepo}{\url{https://anonymous.4open.science/r/AgenticECO}}
\newtcolorbox{promptbox}[1][]{
  enhanced, breakable, width=\linewidth,
  colback=blue!3!white, colframe=black!75,
  colbacktitle=black!80, coltitle=white,
  fonttitle=\bfseries\footnotesize, boxrule=0.4pt, arc=2pt,
  left=6pt, right=6pt, top=4pt, bottom=4pt,
  fontupper=\small\raggedright,
  before skip=0.5\baselineskip, after skip=0.5\baselineskip, #1}
\definecolor{codebg}{rgb}{0.97,0.98,1.00}
\definecolor{codecomment}{rgb}{0.20,0.55,0.30}
\definecolor{codeframe}{rgb}{0.35,0.45,0.65}
\lstdefinestyle{repotree}{
  backgroundcolor=\color{codebg}, basicstyle=\ttfamily\scriptsize,
  commentstyle=\color{codecomment}, morecomment=[l]{\#},
  frame=single, rulecolor=\color{codeframe}, framesep=4pt,
  xleftmargin=4pt, xrightmargin=4pt, breaklines=true,
  breakindent=14pt, columns=fullflexible, keepspaces=true}

\newcommand{\system}{AgenticECO}
\newcommand{\engine}{\text{TaiWei}}
\newcommand{\ecoroute}{\text{EcoRoute}}
\newcommand{\hbt}{HBT}

\title{AgenticECO: An Agentic Framework for ECO on 3D Integrated Circuits}
\author{Shuo Ren, Yaohui Han, Libo Shen, Zhiqiang Jia, Rongliang Fu, Bei Yu, Tsung-Yi Ho}
\affiliations{The Chinese University of Hong Kong}

\begin{document}
\maketitle

\begin{abstract}
As Moore's law slows, the industry is turning to three-dimensional
integration; yet in merged 3D-IC flows, routed designs expose bond-level
defects with no 2D analogue, and post-route engineering change orders (ECO)
remain manual, expertise-bound work.  Worse, the standard
edit-then-fully-reroute practice entangles a repair with router churn, so
a signoff number cannot be attributed to the edit that motivated it.  We
present AgenticECO, an evidence-gated
tool-using agent workflow for 3D-IC ECO on the open-source TaiWei flow,
paired with EcoRoute, a minimal-disturbance ECO-routing layer
that drives the unmodified pinned router so a repair is attributable to its
edit.  Across nine matched natural defect cases under identical budgets,
AgenticECO clears seven versus two for both full reroute and stock repair,
at 0.66\% mean disturbance over cleared cases and zero clock nets
touched, and a
cross-backbone rerun under the same sealed contract clears all nine.
Controlled studies show that the repair moves are necessary under preservation,
that occupancy-aware choice buys legal landings rather than repair
success, and that under tightened clocks minimal disturbance flips accept
versus reject.  Three preregistered visual studies localize the
pixel instrument's edge to contested landing sites, and a preregistered
blind diagnostic exactly restores every held-out injected defect, the
only arm with zero wrong edits.  Every accepted result
passes routing, fresh extraction, max/min timing, DRC, and
structural-equivalence gates.  Code, environment, and per-episode audit
artifacts are released with the paper.
\end{abstract}

\section{Introduction}

AI agents are moving beyond conversational assistance toward long-horizon
engineering tasks, where they invoke specialized tools, inspect
intermediate artifacts, and revise decisions over multiple rounds
\cite{react,toolformer,toollm,agentbench,codeact,reflexion,sweagent,
osworld}, with chip design a prominent recent example \cite{kimik3chip}.

Modern chip design is not performed by a single algorithm: it is a sequence
of specialized tools for synthesis, placement, routing, timing analysis, and
physical verification, each stage producing the design state and evidence
needed by later decisions.  Early work established conversational chip
co-design, domain-adapted models, RTL generation and evaluation, and
tool-feedback repair
\cite{chipchat,chipnemo,llm4eda,verigen,rtllm,gpt4aigchip,autochipfeedback};
recent Agentic EDA systems generate and repair scripts, coordinate design
tools, optimize flows, modify engine code, and assist physical design
\cite{chateda,edaid,mage,jarvis,layoutcopilot,openroadagent,orfsagent,
agenticedasurvey}, vendors ship multi-tool automation
\cite{chipstack,agentengineer,siemensfuse}, and FluxBench benchmarks
agents on full RTL-to-GDS flows \cite{fluxbench2026}.  Together these
efforts position EDA as a
long-horizon, tool-interactive agent task rather than a collection of
isolated model predictions.

\begin{figure}[t]
\centering
\includegraphics[width=.94\columnwidth]{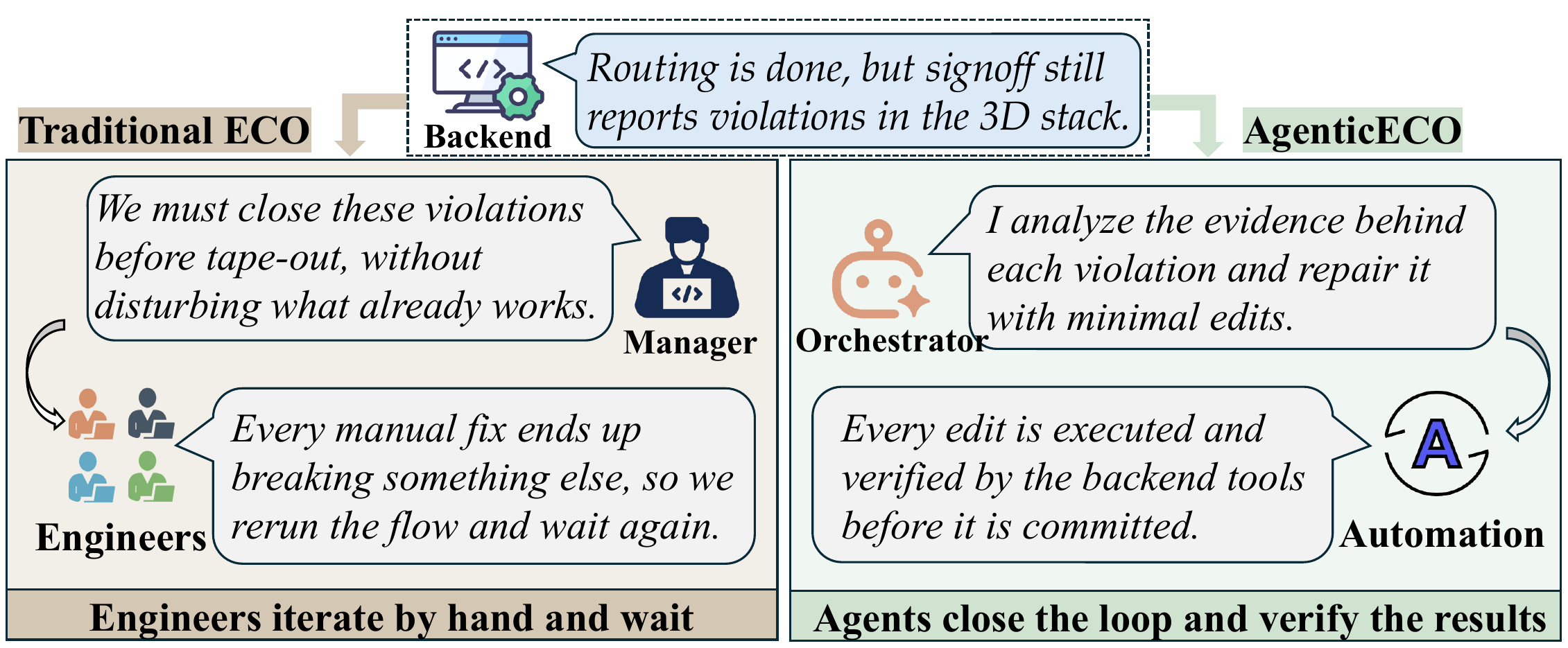}
\caption{Traditional ECO vs.\ \system{}'s verified loop.}
\vspace{-0.5cm}
\label{fig:motivation}
\end{figure}

\begin{figure*}[t]
\centering
\setlength{\tabcolsep}{3pt}
\begin{tabular}{cc}
\includegraphics[height=1.40in]{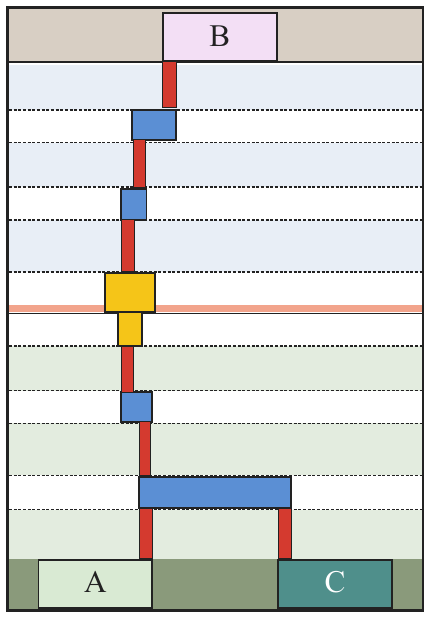} &
\includegraphics[height=1.40in]{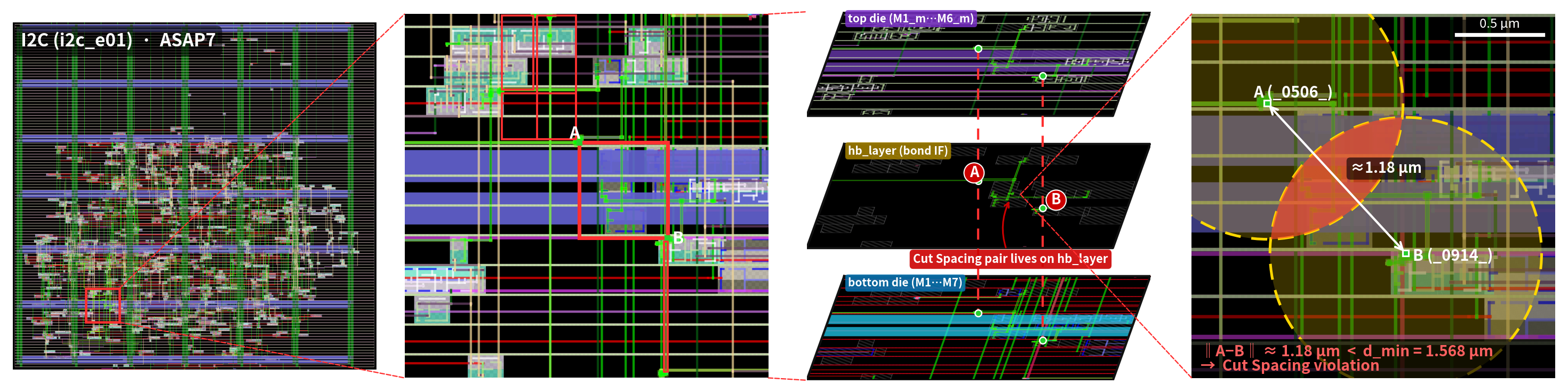} \\
\footnotesize (a) F2F hybrid bonding &
\parbox[t]{5.35in}{\centering\footnotesize (b) routed I2C: merged die,
defect window, per-tier reveal, violating pad pair} \\
\end{tabular}
\caption{The problem we target.  (a) F2F stacking joins dies through
bond-pad pairs (after \citet{zhao2025advancing3d}).  (b) The merged die
looks legal, yet two pads sit at $1.18\,\mu$m under a $1.568\,\mu$m rule
(Section~\ref{sec:prelim}), a cross-tier violation no 2D ECO represents.}
\label{fig:defect}
\end{figure*}

A particularly demanding task arises after a chip has already been placed
and routed.  Engineers use an engineering change order (ECO) to correct
localized physical violations while preserving the valid
implementation \cite{ecobsdrv}.
In practice, ECO is a careful debugging loop
(Figure~\ref{fig:motivation}): an engineer inspects
heterogeneous evidence, forms a repair hypothesis, generates an edit,
performs ECO routing, runs signoff, and interprets the new outcome; a
failed attempt may require a different candidate, additional evidence, or
a different procedure altogether.  The loop is labor intensive and
state-dependent: routing and signoff expose physical consequences
unavailable when the edit was proposed.

This burden grows as geometric scaling nears its limits
\cite{mooreslaw} and design moves into three dimensions.  Hybrid
bonding enables dense vertical connections \cite{hybridbond}, merged
physical-design methodologies represent multiple tiers in a unified,
tier-encoded database \cite{compact2d,pin3d}, and open engines and
benchmarks provide the substrate \cite{openroad,open3dbench}.  Even
after routing, such designs retain cross-tier defects;
Figure~\ref{fig:defect} shows a hybrid-bond spacing violation.

These developments establish the promise of Agentic EDA, but post-route
physical repair requires more than iterative tool use: the system must
reason over evolving execution outcomes.  A placement-legal candidate may
fail after routing, meet geometry but violate timing, or expose a new
violation whose evidence refutes the assumptions behind the candidate,
so resampling a fixed procedure is insufficient.  Language agents refine outputs through self-feedback,
tool-grounded critique, environment-guided search, or learned rules
\cite{selfrefine,critic,lats,toolplanner,automanual,conagents}, but
physical repair requires distinguishing a failed candidate from a refuted
candidate family and revising the executable procedure that generates
later edits.  The closest repair systems act on 2D geometry: learned ECO
in a fixed action space \cite{irawareeco}, or evolved layer-wise skills over
bounded layout crops \cite{evodrc}.  Neither executes routed,
cross-tier-preserving 3D ECOs.  To our knowledge, prior
Agentic EDA systems
\cite{chateda,orfsagent,audopeda,openllmeco,pdagentbench} have not used typed
post-execution artifacts to drive semantic revision of a registered,
deterministic repair procedure under fixed external acceptance gates for
post-route 3D-IC ECO.  Appendix~P surveys related agent,
hardware, and 3D-IC work in full.

We introduce \system{}\footnote{The full code, the pinned environment,
and per-episode audit artifacts are released at \coderepo; Appendix~B documents their organization and reproduction steps.}, an evidence-gated agentic framework for this setting.
The orchestrator accumulates typed evidence from the netlist, layout, violations,
routing, timing, and prior attempts, and uses that evidence to synthesize a
deterministic repair program.  Exact, registered candidate edits come from
that program rather than from free-form model text.  Each candidate is applied
by the \ecoroute{} executor and judged by the independent verifier.  After a
failure, the orchestrator explicitly chooses \text{Advance} for the next
registered candidate, \text{Query} for additional evidence, or
\text{Revise} for a new repair program.

Reliable revision also requires reliable feedback.  \system{} therefore
separates model reasoning from physical execution and acceptance.  The bounded
\ecoroute{} executor applies each edit on a copy of the design while recording the realized
change and its routing disturbance.  An independent verifier evaluates
routing completion, design-rule checks, max/min timing, structural
equivalence, and preservation constraints using tool artifacts only; the
orchestrator may revise its procedure after a failure, but it cannot revise the
acceptance gates.  This separation makes failure evidence attributable,
exact actions replayable, and accepted repairs externally verifiable.

Our contributions are summarized as follows:
\begin{itemize}
  \item To the best of our knowledge, \system{} is the first agentic framework that
  repairs post-route \hbt{}-level DRC violations in 3D integrated circuits.
  \item We formulate the task around its central difficulty: an edit is
  judged only after routing runs, so each verified failure forces a
  choice between the next candidate, more evidence, and a new repair
  procedure.
  \item We design the multi-agent information flow: read-only
  specialists distill the database into typed, provenance-bearing
  evidence, the orchestrator compiles it into a repair program that
  emits every coordinate, and an artifact-only verifier judges each
  localized edit.
  \item Experiments show \system{} repairs defects that full reroute and
  stock repair leave standing, at a fraction of their disturbance with
  zero clock touches, and overturns the single agent's wrong
  infeasibility verdicts.
\end{itemize}

\section{Preliminaries}
\label{sec:prelim}

\begin{figure*}[!t]
  \centering
  \includegraphics[width=.98\textwidth]{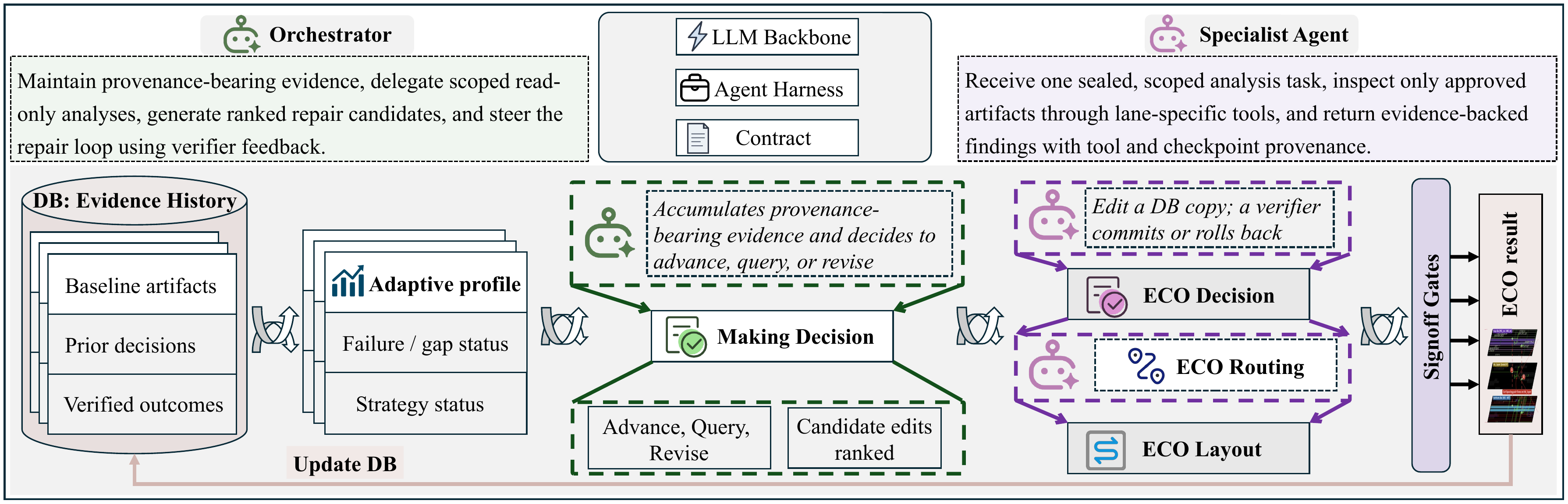}
  \caption{\system{} framework.  Read-only specialist agents return
  evidence-backed findings; the orchestrator maintains the
  provenance-bearing evidence history with its adaptive profile and
  decides \text{Advance}, \text{Query}, or \text{Revise}; the registered
  repair program emits ranked candidate edits; \ecoroute{} executes each
  edit on a database copy; and the independent verifier's signoff gates
  commit verified $X'$ or roll back.}
  \vspace{-0.5cm}
  \label{fig:agenticeco-flow}
\end{figure*}

\subsection{The Physical State and the 3D Defect}
\label{subsec:flow}

We introduce only the physical-design objects needed to define the agent's
task.  A face-to-face two-tier design (Figure~\ref{fig:defect}a) is represented
in one merged database \cite{compact2d,pin3d}.  Its routed state is
\begin{equation}
  X=(\mathcal I,\mathcal N,t,p,R),
  \label{eq:ckpt}
\end{equation}
where $\mathcal I$ and $\mathcal N$ are instances and nets, $t(i)$ and $p_i$
give an instance's tier and location, and $R$ is routed geometry.  Every repair
starts from a frozen checkpoint; $X'$ denotes the resulting state.

For the post-route task studied here, connectivity and tier assignment are
frozen.  Placement has chosen $p$ using cell-body constraints, whereas routing
draws wires between the resulting pins and determines where a cross-tier wire
crosses the bonding interface:
\begin{equation}
  \bigl(R,\{B_n\}_{n\in\mathcal N_\times}\bigr)
  =\operatorname{Route}(\mathcal I,\mathcal N,t,p;\theta_R).
  \label{eq:route}
\end{equation}
Here $B_n\subset\mathbb R^2$ is the final \hbt{} geometry and $\theta_R$ is
the frozen router configuration.  Thus legal cell placement alone does not
guarantee legal spacing between terminals created later.

Let $\mathcal N_\times$ be the cross-tier nets and $s_B$ the minimum terminal
spacing.  The target bond-level defects are
\begin{equation}
  \mathcal D_B(X)=
  \bigl\{\{m,n\}\subseteq\mathcal N_\times:
  \operatorname{sep}_B(B_m,B_n)<s_B\bigr\},
  \label{eq:hbt-conflict}
\end{equation}
where $\operatorname{sep}_B$ is the backend's rule-specific geometry spacing.
In our technology setup an \hbt{} is $0.032\,\mu$m wide whereas
$s_B=1.568\,\mu$m, so legal cell bodies can produce violating terminals later
(Figure~\ref{fig:defect}b).  The merged backend folds both tiers and the
bond layer into one routing stack.

\subsection{The Agent's Information Boundary}
\label{subsec:agent-view}

From an agent's perspective, the routed state $X$ of Eq.~\eqref{eq:ckpt} is
environment state, not prompt text.  The
full-chip database and its reports exceed a practical context window, while
the answer to a repair question often resides in a small region or tool
result.  The task therefore provides three distinct information roles.  A
fixed contract
$\Gamma=(\text{schema},\text{rules},\text{edit vocabulary},
\text{acceptance gates})$ guides the
agent without identifying a successful repair.  A tool query $q_t$ reveals a
scoped observation $o_t=Q_{q_t}(X_t)$.  Finally, feedback $f_t$ becomes
available only after an edit changes the physical state.  Thus the agent acts
from the bounded history
\begin{equation}
  \begin{aligned}
  h_t=\bigl(\Gamma,\{(q_j,Q_{q_j}(X_j))\}_{j\le t},
  \{f_j\}_{j<t}\bigr),
  \end{aligned}
  \label{eq:observation}
\end{equation}
not from all of $X$.  The research problem is deciding which $q_t$ to issue
and how to revise a repair after $f_t$ contradicts its assumptions;
Section~\ref{subsec:generate} specifies how \system{} turns database
queries into model-readable evidence.

\subsection{ECO as Partially Observed Repair}
\label{subsec:blindspot}

In our repair interface, the agent influences the \hbt{} terminal geometry
$B_n$ of Eq.~\eqref{eq:route} only indirectly, by relocating a small same-tier
set of endpoint instances,
\begin{equation}
  a=\{i\mapsto p_i' : i\in\mathcal I_a\},
  \qquad t'(i)=t(i),\quad |\mathcal I_a|\ll|\mathcal I|,
  \label{eq:action}
\end{equation}
and the \ecoroute{} executor reroutes invalidated nets to produce
$X'=T(X,a)$.  Transition
$T$ has no closed form: route feasibility and collateral violations appear
only after execution.

Localized repair should also preserve the already-valid implementation.
Using a canonical geometry fingerprint $H_n(X)$ for each net, we measure
\begin{equation}
  \Delta_{\mathrm{net}}(X,X')=
  \bigl|\{n\in\mathcal N:H_n(X')\ne H_n(X)\}\bigr|;
  \label{eq:disturbance}
\end{equation}
small disturbance suppresses unrelated routing churn and keeps feedback
attributable to the requested edit.  The task is thus
\begin{equation}
  \begin{aligned}
  \min_a\quad &\Delta_{\mathrm{net}}(X,T(X,a))\\
  \mathrm{s.t.}\quad
    &\mathcal D_B(T(X,a))=\emptyset,\\
    &\mathcal V(X,T(X,a))=\text{pass},
  \end{aligned}
  \label{eq:problem}
\end{equation}
where the fixed verifier $\mathcal V$ checks the declared routing, DRC,
realized-edit, connectivity, and preservation criteria.  The model neither
computes $T$ nor declares \text{pass}: the agentic problem is deciding
what to inspect and how to revise a repair procedure when execution
refutes its assumptions.

\section{Methodology}
\label{sec:method}

\subsection{AgenticECO Framework}
\label{subsec:framework}

The objective of Eq.~\eqref{eq:problem} is observable only after a physical
edit executes.  \system{} therefore repeats four named stages: query scoped
evidence, generate a repair program, execute one registered edit, and verify
its artifacts (Figure~\ref{fig:agenticeco-flow}).  Reasoning may expand the
evidence available to the orchestrator, but only an edit $a$ changes the
database through $T(X,a)$.

The orchestrator $\mathcal A$ maintains an evidence set $E$, initialized
with baseline tool artifacts and only ever grown under the fixed contract
$\Gamma$ of Section~\ref{subsec:agent-view}.  An independent verifier
$\mathcal V$ closes the loop.  One round is
\begin{equation}
  \label{eq:loop}
  \begin{aligned}
    E\;&\leftarrow\;E\cup\{Q_q(X)\}
    &&\text{query}\\
    g\;&\sim p_{\mathcal A}(\cdot\mid\Gamma,E),\quad
    (\mathcal C,\prec)=g(X)
    &&\text{generate}\\
    X'\;&=\;\text{EcoRoute}(X,\,a)
    &&\text{execute}\\
    V\;&=\;\mathcal V(X,X')\in\{\text{pass},\text{fail}\}
    &&\text{verify}
  \end{aligned}
\end{equation}
Here the orchestrator selects $q$ from the queries approved by $\Gamma$, and $g$
is a deterministic repair program synthesized from $\Gamma$ and accumulated
evidence $E$.  Through approved APIs, $g$ maps the tool-retained state to ranked candidates
$(\mathcal C,\prec)$ without exposing $X$ to the model.  The highest-ranked
untried candidate executes; a \text{pass} commits $X'$, whereas a
\text{fail} rolls back to $X$ and appends its artifacts to $E$.  The new
evidence then determines whether the orchestrator selects \text{Advance}
within $\mathcal C$, \text{Query} for more evidence, or \text{Revise} to
replace $g$.  This failure-conditioned procedure revision,
rather than repeated sampling from a fixed procedure, is the central control
mechanism.

Algorithm~\ref{alg:loop} implements this control loop under budget $K$.

\begin{algorithm}[t]
\caption{\system{} closed-loop repair}
\label{alg:loop}
\small
\setlength{\hsize}{0.95\linewidth}
\KwIn{Checkpoint $X$, contract $\Gamma$, verification budget $K$}
\KwOut{Accepted state $X'$, or verified failure with audit trail}
$E = \text{InitialEvidence}(X)$;\quad $k = 0$\;
\While{$k < K$}{
  $q = \text{SelectQuery}_{\mathcal A}(\Gamma,E)$\;
  $E = E \cup \{Q_q(X)\}$ \tcp*{query, Eq.~\eqref{eq:loop}}
  $g \sim p_{\mathcal A}(\cdot\mid\Gamma,E)$ \tcp*{generate, \S\ref{subsec:generate}}
  $(\mathcal C,\prec) = \text{Register}(g, X)$ \tcp*{ranking Eq.~\eqref{eq:coupled}}
  \While{$\text{Untried}(\mathcal C)\ne\emptyset$ \textup{\textbf{and}} $k<K$}{
    $a = \text{FirstUntried}(\mathcal C,\prec)$\;
    $k = k+1$\;
    $X' = \text{EcoRoute}(X, a)$ \tcp*{execute, Eq.~\eqref{eq:victim}}
    $V = \mathcal V(X, X')$ \tcp*{verify, \S\ref{subsec:execute}}
    \lIf{$V=\text{pass}$}{\Return{$X'$}}
    $E = E \cup \text{Artifacts}(X')$;\quad roll back to $X$\;
    \Repeat{$d\ne\text{Query}$}{
      $d = \text{Decide}_{\mathcal A}(\Gamma,E,\mathcal C)$\;
      \lIf{$d=\text{Query}$}{$q=\text{SelectQuery}_{\mathcal A}(\Gamma,E)$;
        $E=E\cup\{Q_q(X)\}$}
    }
    \lIf{$d=\text{Revise}$}{\textbf{break} \tcp*[f]{regenerate $g$}}
  }
}
\Return{\textup{verified failure with the full audit trail}}\;
\end{algorithm}

After every failure the orchestrator selects an explicit control action:
\text{Query} folds freshly acquired evidence back into the same
deliberation, so the loop leaves it only with a commitment, and
\text{Advance} to the next registered candidate is that ranking
re-affirmed under the failure's evidence, not a default.
\text{Revise} declares the whole family refuted, exits the inner loop,
and regenerates $g$ from the enlarged evidence set.

\subsection{Specialist Queries and Repair-Program Generation}
\label{subsec:generate}

Generation converts heterogeneous physical evidence into database-consistent
candidates.  \system{} delegates bounded, read-only queries and compiles the
returned evidence into the deterministic program $g$.

\paragraph{Specialist evidence queries.}
The orchestrator $\mathcal A$ never serializes $X$ into a context window.
Instead, each specialist agent receives a contract naming its question,
tools, and permitted
artifacts.  Tool adapters reduce database objects, reports, and rendered
windows to typed records containing query scope, normalized payload, producing
tool, and checkpoint provenance.  Specialist agents cover source and netlist
semantics, local layout, the \hbt{} conflict graph of
Eq.~\eqref{eq:hbt-conflict}, and router or violation reports.  Every returned
claim cites its artifacts, and every measurement originates from a tool.
Rendered-layout records (\text{V-Pix}) encode local occupancy only, never a
downstream outcome.  Concurrent views supply complementary evidence and diverse
candidate hypotheses without exposing the full database.

\paragraph{Adaptive edit generation.}
The orchestrator merges these records into a coordinate-free repair specification:
target violations, movable instances, protected objects, and the fixed gates
inherited from $\Gamma$.  It then synthesizes $g$, which queries legal sites and emits ranked
edits.  Because $g$ is conditioned on all of $E$, failure artifacts can exclude
a refuted family, narrow the region, or replace single relocations with tuples
for coupled conflicts.  Pre-execution tool records define deterministic
residual-conflict and occupancy scores $c(a)$ and $o(a)$:
\begin{equation}
  a\prec a'\iff c(a)<c(a')\ \lor\
  \bigl(c(a)=c(a')\wedge o(a)<o(a')\bigr).
  \label{eq:coupled}
\end{equation}
Remaining ties use canonical instance and site identifiers.  These scores
order feasible hypotheses; localized execution constrains disturbance and the
verifier measures it, without claiming a global optimum of Eq.~\eqref{eq:problem}.

\paragraph{The model writes the program; the program writes the edits.}
Sampling stops at $g$.  Registration fixes every coordinate, ranking, scope,
and inherited gate before execution, making the round deterministic and
replayable and preventing post-hoc reordering or softened acceptance.

\subsection{Localized ECO Execution and Verification}
\label{subsec:execute}

The execution stage applies a proposed edit so that the observed outcome is
attributable to the edit; the verification stage judges the resulting
artifacts independently of the model.  Invoking backend routing without an explicit preservation
scope can change nets the edit never touched, conflating the edit with
unrelated routing churn.

\paragraph{Localized execution.}
\system{} executes every edit through the \ecoroute{} executor,
which drives the unmodified pinned router on a copy of the design.
For an edit $a$ on instances $\mathcal I_a$, the initial reroute set
contains exactly their incident nets,
\begin{equation}
  \mathcal R_0=
  \{n\in\mathcal N:n\text{ is incident to some }i\in\mathcal I_a\},
  \label{eq:victim}
\end{equation}
and every other net is frozen.  Let $\mathcal R$ be the reroute set of a
routing round, $X'$ its output, and $\mathcal D$ the full DRC violation set.
The round succeeds only if
\begin{equation}
  \begin{aligned}
    &\forall n\in\mathcal R:\ n\text{ is routed in }X',\\
    &\forall n\in\mathcal N\setminus\mathcal R:\ H_n(X')=H_n(X),\\
    &\mathcal D(X')\setminus\mathcal D(X)=\emptyset.
  \end{aligned}
  \label{eq:route-invariants}
\end{equation}
The second condition uses the fingerprints of \eqref{eq:disturbance} to certify
that every net outside $\mathcal R$ is unchanged; the complete ledger then
computes $\Delta_{\mathrm{net}}$ across both tiers and the bond layer.  If a new violation implicates previously
frozen signal nets $\mathcal U_k$, they are admitted in a bounded
escalation,
\begin{equation}
  \mathcal R_{k+1}=\mathcal R_k\cup\mathcal U_k,\qquad k\le3,
  \label{eq:escalate}
\end{equation}
and clock nets are never admitted.

\paragraph{Verification outside the model.}
The verdict is produced by $\mathcal V$, a frozen deterministic verifier
whose source is archived alongside every verdict it emits.  It compares the
baseline and candidate artifacts (router completion, DRC reports, the
structural-connectivity check, 3D legality, and the
task-specific target) against the gates registered before execution, and
it consumes no model text: no orchestrator or specialist agent statement can enter
the judgment.  What enters the record is the realized edit, recovered from the
output database rather than from the request.
Execution is copy-on-write, so a \text{fail} costs nothing but its
evidence.

\paragraph{Verification-guided refinement.}
The verifier accepts or rolls back; it never optimizes.  Improvement
happens in the next round, where the residual conflict
set $\mathcal D_B(X')$, the router log, and the realized edit enter $E$, and
the orchestrator decides at the decision point of Algorithm~\ref{alg:loop}
whether to \text{Advance}, \text{Query}, or \text{Revise}.
When the evidence shows the objective unreachable within the edit space,
the loop returns a verified failure backed by the same artifacts as
a success.  Every delegation contract, evidence artifact, registered
program, realized edit, and verdict joins an audit trail, so a repair can
be replayed and audited.

\section{Experiments}
\label{sec:exp}

\subsection{Setup}

All experiments run on one workstation: Ubuntu 22.04, two Xeon Gold
6426Y, 251\,GiB RAM.  The backend is the public, unmodified \engine{}
flow, pinned end to end; exact commits are listed in Appendix~A, and the
agent contracts, tool registry, ledger, worked trajectory, renderer,
glossary, clean-room protocol, and scope limits in Appendices~C--I and~O.  The technology is
the ASAP7 predictive PDK; the three designs are GCD, a macro-free I2C
peripheral, and a UART.  The cases are nine natural post-route
cut-spacing defects (three per design), surfaced by HBT-density and
utilization sweeps rather than injection, frozen with their baseline
signoff before any method runs.  Agentic methods run on two frozen
backbones: Claude Opus 4.8 under the Claude Code harness, and GPT-5.6
under the Codex harness.  Every method runs each case once under
an identical $\le\!40$-call, $\le\!3$-signoff budget; a repair must
reach zero DRC (throughout, a DRC count is the number of design-rule
violations the pinned full-design check reports), produce the
disturbance ledger of
\eqref{eq:disturbance}, preserve setup WNS (or leave it nonnegative),
avoid worse TNS, and preserve structural equivalence.  Cut-spacing
violations threaten manufacturability, not connectivity, so timing
enters as this non-regression gate, not an outcome axis.

\subsection{Surgical Repair and Attribution}

\begin{table*}[t!]
\centering
\caption{Matched end-to-end repair quality on nine natural post-route
hybrid-bond defect cases (three per design: 1/2/3): per-design DRC and changed-net
triplets, plus nine-case robustness aggregates.}
\label{tab:main}
\setlength{\tabcolsep}{1.8pt}
\scriptsize
\resizebox{\linewidth}{!}{%
\begin{threeparttable}
\renewcommand{\arraystretch}{1.10}
\begin{tabular}{l ccc ccc ccc ccccccccc}
\toprule
& \multicolumn{3}{c}{\textbf{GCD} (${\approx}600$ nets/case)}
& \multicolumn{3}{c}{\textbf{I2C} (${\approx}1{,}250$ nets/case)}
& \multicolumn{3}{c}{\textbf{UART} (${\approx}900$ nets/case)}
& \multicolumn{9}{c}{\textbf{All nine cases}} \\
\cmidrule(lr){2-4}\cmidrule(lr){5-7}\cmidrule(lr){8-10}
\cmidrule(lr){11-19}
\textbf{Method}
& \shortstack{Full DRC\\out (1/2/3)}$\downarrow$
& \shortstack{$\Delta$Nets\\(1/2/3)}$\downarrow$
& \shortstack{DRC0\\cases}$\uparrow$
& \shortstack{Full DRC\\out (1/2/3)}$\downarrow$
& \shortstack{$\Delta$Nets\\(1/2/3)}$\downarrow$
& \shortstack{DRC0\\cases}$\uparrow$
& \shortstack{Full DRC\\out (1/2/3)}$\downarrow$
& \shortstack{$\Delta$Nets\\(1/2/3)}$\downarrow$
& \shortstack{DRC0\\cases}$\uparrow$
& \shortstack{$\Sigma$DRC\\in$\rightarrow$out}$\downarrow$
& \shortstack{DRC\\reduction}$\uparrow$
& \shortstack{Improved\\cases}$\uparrow$
& \shortstack{DRC0\\cases}$\uparrow$
& \shortstack{All gates\\pass}$\uparrow$
& \shortstack{Mean $\Delta$Nets\\(\%)}$\downarrow$
& \shortstack{Clock\\touched}$\downarrow$
& \shortstack{HBT topology\\preserved}$\uparrow$
& \shortstack{Worst residual\\DRC}$\downarrow$ \\
\midrule
No repair
& 4/6/26 & 0/0/0 & 0/3
& 5/23/6 & 0/0/0 & 0/3
& 9/5/10 & 0/0/0 & 0/3
& 94$\rightarrow$94 & 0.0\% & 0/9 & 0/9 & 0/9 & 0.00 & 0 & 9/9 & 26 \\
Full reroute
& 20/4/47 & 337/331/352 & 0/3
& 0/14/0 & 707/678/699 & 2/3
& 14/6/11 & 508/501/518 & 0/3
& 94$\rightarrow$116 & $-23.4$\% & 4/9 & 2/9 & 2/9 & 55.85 & 98 & 4/9 & 47 \\
Stock repair
& 0/6/27 & 5/3/14 & 1/3
& 1/8/0 & 5/15/5 & 1/3
& 6/5/10 & 8/1/3 & 0/3
& 94$\rightarrow$63 & 33.0\% & 5/9 & 2/9 & 1/9 & 0.77 & 2 & 7/9 & 27 \\
Single agent (Opus 4.8)
& 0/4/26 & 3/9/0 & 1/3
& 0/0/0 & 6/12/5 & 3/3
& 5/0/0 & 8/2/5 & 2/3
& 94$\rightarrow$35 & 62.8\% & 8/9 & 6/9 & \textbf{5/9}
& \textbf{0.61} & \textbf{0} & \textbf{9/9} & 26 \\
\textbf{AgenticECO (Opus 4.8)}
& \textbf{0/0/16} & 3/8/20 & \textbf{2/3}
& \textbf{0/0/0} & \textbf{3/12/4} & \textbf{3/3}
& \textbf{1/0/0} & 17/2/10 & \textbf{2/3}
& \textbf{94$\rightarrow$17} & \textbf{81.9\%} & \textbf{9/9}
& \textbf{7/9} & \textbf{5/9} & 1.08 & \textbf{0} & \textbf{9/9} & \textbf{16} \\
\midrule
Single agent (GPT-5.6)
& 0/0/0 & 3/10/16 & 3/3
& 0/0/0 & 6/12/5 & 3/3
& 0/0/0 & 8/4/10 & 3/3
& 94$\rightarrow$0 & 100.0\% & 9/9 & 9/9 & 6/9 & 1.00 & 0 & 8/9\tnote{*} & 0 \\
AgenticECO (GPT-5.6)
& 0/0/0 & 5/1/2 & 3/3
& 0/0/0 & 3/12/1 & 3/3
& 0/0/0 & 6/4/2 & 3/3
& 94$\rightarrow$0 & 100.0\% & 9/9 & 9/9 & 7/9 & 0.44 & 0 & 9/9 & 0 \\
\bottomrule
\end{tabular}
\begin{tablenotes}[flushleft]
\footnotesize
\item[] Frozen backend, matched budgets; bolding compares the Opus 4.8
block.  Bottom block reruns the same sealed contract on GPT-5.6,
independently re-verified.  Equivalence is structural (no formal LEC).
*: exceeded the call budget; moved one bond via.
\end{tablenotes}
\end{threeparttable}%
}
\end{table*}

\begin{figure}[t]
\centering
\begin{minipage}[t]{0.32\columnwidth}
  \centering
  \includegraphics[width=\linewidth]{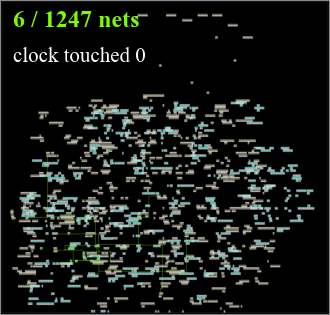}\\[1pt]
  {\scriptsize (a) \ecoroute{}, 3 moves}
\end{minipage}\hfill
\begin{minipage}[t]{0.32\columnwidth}
  \centering
  \includegraphics[width=\linewidth]{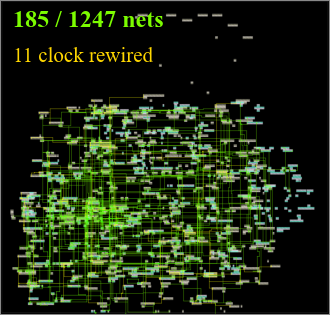}\\[1pt]
  {\scriptsize (b) full reroute, 3 moves}
\end{minipage}\hfill
\begin{minipage}[t]{0.32\columnwidth}
  \centering
  \includegraphics[width=\linewidth]{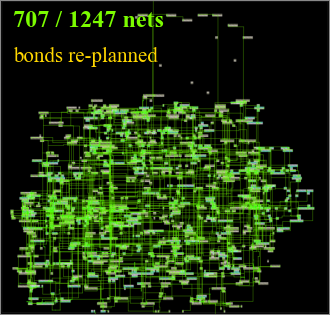}\\[1pt]
  {\scriptsize (c) full reroute, no moves}
\end{minipage}
\caption{Three DRC0 repairs of one I2C defect at identical full-die scale:
lit area (clock in yellow) is the repair budget.}
\vspace{-0.5cm}
\label{fig:footprint}
\end{figure}

\paragraph{A deterministic baseline trichotomy.}
Figure~\ref{fig:footprint} makes the disturbance contrast concrete: all three
I2C runs reach DRC0, but \ecoroute{} touches only 6/1247 nets and no clocks
(a), versus full reroute's 185--707 nets, 11 rewired clocks, or replanned
bonds (b,c).
Table~\ref{tab:main} compares the methods.  Full reroute, a whole-design
rip-up, is a blunt hammer: it clears only 2/9 cases, rewires 55.9\% of
nets on average, churns 98 clock nets, and silently rewrites the
hybrid-bond topology on 5/9 cases.  Stock repair, the backend's built-in
repair pass, is nearly free (0.8\% mean
disturbance, two clock nets) but clears only 2/9 and passes every gate on
one.  Between a churning hammer and a scalpel too weak to cut, deliberate
repair moves are necessary.

\paragraph{\system{} dominates the baselines.}
\system{} clears DRC on \textbf{7/9} cases against full reroute's and stock
repair's 2/9, at 0.66\% mean disturbance over cleared cases, zero clock
nets, and the hybrid-bond via count preserved on all
nine: stock-repair disturbance with far higher clearing power.  The two hardest cases, both uncleared, are GCD-3 (an I/O-pinned bond
site outside the edit space) and UART-1 (Figure~\ref{fig:case});
even there \system{} makes the most DRC
progress in the matched block ($26\!\to\!16$, $9\!\to\!1$).  Its only other gate
losses are sub-picosecond $\Delta$-timing trips on GCD's already-negative
baseline.

\paragraph{Versus a single agent.}
The single-agent control is the same backbone dropped into the same
\system{} environment, keeping the sealed episode, the seven frozen
tool wrappers including the \ecoroute{} executor, the budget, and the
independent verifier; it gives up only delegation, orchestration, and
preregistered revision.  A bare coding agent without this surface
cannot drive the pinned engine at all, so the comparison isolates
agentic organization, not tooling.  Over the shared cases \system{} wins five, ties three, and loses one
(UART-3, 10 disturbed nets versus 5).  But the single agent also reached two wrong infeasibility verdicts, declaring
GCD-2 unresolvable at DRC\,4 and surrendering GCD-3 at DRC\,26 (cases
\system{} improved to 0 and 16), each after only ${\sim}4$ of its 40
tool calls: adaptive orchestration converts unused budget into reached
repairs.

\paragraph{Backbones bound the search, not the contract.}
Rerun with a GPT-5.6 backbone under the identical sealed contract, both
agent arms clear all nine cases at DRC0 (Table~\ref{tab:main}, bottom
block; per-cell in Appendix~J), \system{} at 0.44\% mean disturbance against its single-agent
control's 1.00\%, with zero clock nets touched: the Opus 4.8 residuals are
search floors, not physical ones, and the contract transfers intact.

\begin{figure}[t]
\centering
\includegraphics[width=.98\columnwidth]{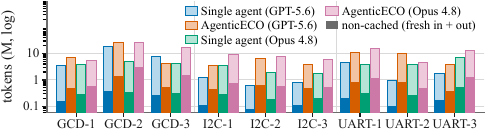}\\[-3.0pt]
{\footnotesize (a) tokens per episode; saturated base is the non-cached part}\\[1.0pt]
\includegraphics[width=.98\columnwidth]{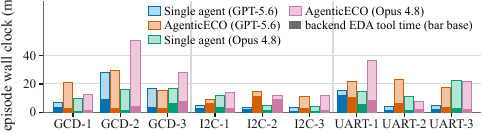}\\[-3.0pt]
{\footnotesize (b) wall clock, backend tool time at the bar base}\\[1.0pt]
\includegraphics[width=.98\columnwidth]{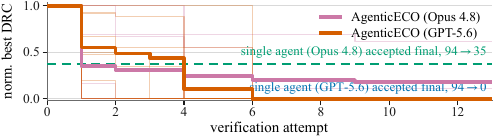}\\[-3.0pt]
{\footnotesize (c) best verified DRC by attempt (thin: per episode; bold: sum)}
\caption{Cost and convergence of the four agentic arms of
Table~\ref{tab:main}; both \system{} arms undercut the Opus 4.8 single-agent
accepted final within four attempts.  Nothing is estimated.}
\label{fig:cost}
\end{figure}

\begin{table*}[t!]
\centering
\caption{UART planner-by-executor replay on one development-exposed case:
two committed placement plans replayed through both routing executors,
separating plan provenance from physical execution.}
\label{tab:system}
\setlength{\tabcolsep}{1.7pt}
\scriptsize

\resizebox{\textwidth}{!}{%
\begin{threeparttable}
\renewcommand{\arraystretch}{1.08}
\begin{tabular}{ll l ccc ccc ccc ccccc}
\toprule
\multicolumn{3}{c}{\textbf{Configuration}}
& \multicolumn{3}{c}{\textbf{Repair}}
& \multicolumn{3}{c}{\textbf{Locality}}
& \multicolumn{3}{c}{\textbf{PPA}}
& \multicolumn{5}{c}{\textbf{Measured cost / source trace}} \\
\cmidrule(lr){1-3}\cmidrule(lr){4-6}\cmidrule(lr){7-9}
\cmidrule(lr){10-12}\cmidrule(lr){13-17}
\textbf{Plan source}
& \textbf{Executor}
& \shortstack{Route\\scope}
& \shortstack{Fresh full DRC\\in$\rightarrow$out}$\downarrow$
& \shortstack{Removed\\count / \%}$\uparrow$
& \shortstack{Moved\\cells}$\downarrow$
& \shortstack{$\Delta$Nets / kept\\count / \%}$\downarrow$
& \shortstack{Clock\\touched}$\downarrow$
& \shortstack{HBT\\vias}
& \shortstack{WNS max/min\\(ps)}$\uparrow$
& \shortstack{Power\\(mW)}$\downarrow$
& \shortstack{WL\\($\mu$m)}$\downarrow$
& \shortstack{Route\\rounds}$\downarrow$
& \shortstack{EDA\\wall (s)}$\downarrow$
& \shortstack{Platform\\calls}$\downarrow$
& \shortstack{Total\\tokens}$\downarrow$
& \shortstack{Agent\\wall (s)}$\downarrow$ \\
\midrule
Single
& Stock & all nets
& 15$\rightarrow$4 & 11/15 (73.3) & 3/850
& 77/881 / 91.26 & 2/10 & 63/63
& 9.314/4.159 & 1.7909 & 4806.2
& 1 & 84 & \textbf{19} & \textbf{1.325M} & \textbf{396.0} \\
Single\tnote{*}
& \ecoroute{} & 6 signal
& 15$\rightarrow$4 & 11/15 (73.3) & 3/850
& \textbf{6/881 / 99.32} & \textbf{0/10} & 63/63
& \textbf{9.314/4.159} & 1.7913 & 4803.0
& 1 & \textbf{31} & 19 & 1.325M & 396.0 \\
1+2\tnote{*}
& Stock & all nets
& 15$\rightarrow$9 & 6/15 (40.0) & 3/850
& 43/881 / 95.12 & \textbf{0/10} & 63/63
& 9.314/4.157 & 1.7913 & \textbf{4797.7}
& 1 & 78 & 66 & 3.769M & 633.5 \\
\textbf{1+2}
& \textbf{\ecoroute{}} & \textbf{6 signal}
& \textbf{15$\rightarrow$4} & \textbf{11/15 (73.3)} & \textbf{3/850}
& \textbf{6/881 / 99.32} & \textbf{0/10} & \textbf{63/63}
& \textbf{9.314/4.159} & \textbf{1.7913} & 4801.5
& \textbf{1} & 57 & 66 & 3.769M & 633.5 \\
\bottomrule
\end{tabular}
\begin{tablenotes}[flushleft]
\footnotesize
\item[*] counterfactual replay: the committed placement is reused and only
the executor changes (costs repeat for provenance).  DRC is a fresh
full-design recheck; TNS stays $0$.  A one-case mechanism diagnostic, not
an efficacy claim.
\end{tablenotes}
\end{threeparttable}%
}
\end{table*}

\paragraph{Ablations locate the mechanism.}
The I2C/UART component ablations (figure and raw table in Appendix~K)
localize the damage: outside the executor ablation every I2C case
clears, and the removed factor decides which contested UART case
survives.
Removing \ecoroute{} collapses \system{} onto the hammer: on I2C-1
its accepted layout is byte-identical to the no-agent full-reroute
reference, and over six cases it is indistinguishable from full
reroute (55.9\% disturbance, 82 clock nets).  Removing vision
(\text{V-Pix}) is asymmetric: it matches \system{} on all I2C cases and grid-clean UART-3
but fails on exactly the two contested UART sites, at DRC\,6 and DRC\,5,
the latter with a false out-of-scope verdict; pixels lose precisely the
contested landings, as the visual studies predict.
Removing delegation preserves quality on 5/6 cases but fails the
hardest, UART-1 (DRC\,6 versus 1): one candidate line cannot diversify
enough.  Table~\ref{tab:system} complements these ablations with a
development-exposed UART planner-by-executor replay, mechanism and cost
evidence rather than a causal success-rate claim.

\paragraph{Delegation width and experience.}
Shrinking delegation width monotonically raises residual damage: total
residual DRC 1 at widths up to three, 6 for serial, and 15 for root-only;
the root-only residual falls entirely on the UART cases: width is spent
where defects are contested.
A predeclared eight-rule generic playbook
is double-edged: the experience arm alone fully clears the hardest case
(UART-1, $9\!\to\!0$), yet it leaves DRC\,6 on UART-3 and only ties the
serial ablation's total ($58\!\to\!6$): experience relocates which case
is hard; width buys robustness.

\paragraph{What the search costs, and what it buys.}
Figure~\ref{fig:cost}a,b measure tokens and wall clock on all four
agentic arms; Figure~\ref{fig:cost}c shows what they buy: summed
best-verified DRC undercuts the Opus 4.8 single agent's final on the
first (Opus) and fourth (GPT-5.6) attempts, reaching zero on GPT-5.6 by
attempt six.  Fan-out costs
two to three times the single-agent control's tokens (76.2M versus 39.0M on GPT-5.6,
104.5M versus 35.3M on Opus 4.8), but 93.2\% and 91.4\% of that input is
cache reads, and part of the single agent's smaller bill is surrender:
its wrong infeasibility verdicts each spent ${\sim}4$ of 40 calls,
leaving the violations standing.  The extra tokens buy the repairs it
never reached ($94\!\to\!17$ versus $94\!\to\!35$ on Opus 4.8) and half
its disturbance at DRC0 on GPT-5.6.  Routing and signoff occupy 47/163 minutes (GPT-5.6) and 38/194
(Opus 4.8); agent time dominates.

\paragraph{Case study: anatomy of one repair.}
Figure~\ref{fig:case} shows the hardest matched case, UART-1: nine of
22 hybrid-bond anchors in a $1.6\times1.1\,\mu$m cluster inside the
$8\times8\,\mu$m window of panel a, with $+18$\,ps slack.  Round 1 tries the cheapest
family, rerouting alone, and refutes it: freed bond vias re-collide at
the same crowded sites (DRC $9\!\to\!6$ and $9\!\to\!12$).
Round 2 escalates to move-plus-reroute and plateaus at DRC\,5, but its
signoff returns WNS bit-identical to baseline: the defect nets sit off
the critical path, so displacement is free.  Round 3 tests the hammer once and rejects it as strictly worse
(DRC\,14, 508/907 nets, nine of ten clock nets; panel c).  Round 4 spends the proven-free timing, spreading eight anchors to a
$\ge\!2\,\mu$m pitch, and is committed as the best verified state:
DRC $9\!\to\!1$ at 17/907 nets and zero clock touches (panels d--f).  Round 5 probes the residual pair, finds it
invariant under every family tried, and reports ``DRC $=$ 1 is
the hard floor'': a verified floor, not a cosmetic zero (the
cross-backbone rerun later cleared it), in 13 of 40 tool calls.
The single agent on the same sealed case stopped at DRC\,5,
declining any cell move as ``a gamble \ldots\ judged
unjustified''; the move it feared, \system{} had proven
timing-free and committed.

\begin{figure}[t]
\centering
\setlength{\tabcolsep}{1.0pt}
\renewcommand{\arraystretch}{0.9}
\begin{tabular}{ccc}
\includegraphics[width=.324\columnwidth]{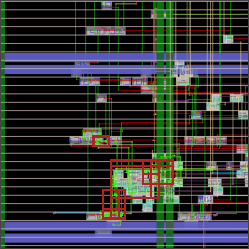} &
\includegraphics[width=.324\columnwidth]{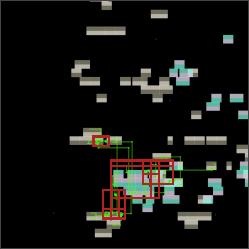} &
\includegraphics[width=.324\columnwidth]{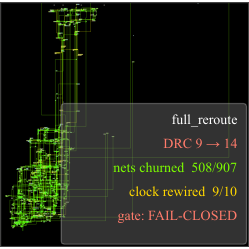} \\
\footnotesize (a) defect window, & \footnotesize (b) bond-pad layer, &
\footnotesize (c) rejected full \\[-2.2pt]
\footnotesize baseline: DRC 9 & \footnotesize victim nets lit &
\footnotesize reroute (clock hit) \\
\addlinespace[2.5pt]
\includegraphics[width=.324\columnwidth]{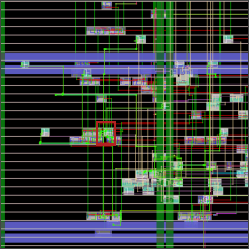} &
\includegraphics[width=.324\columnwidth]{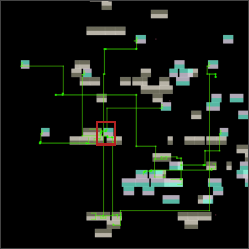} &
\includegraphics[width=.324\columnwidth]{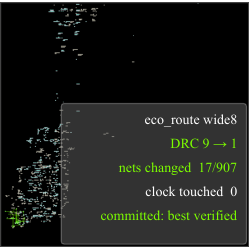} \\
\footnotesize (d) same window, & \footnotesize (e) pads spread to &
\footnotesize (f) committed \\[-2.2pt]
\footnotesize repaired: DRC 1 & \footnotesize legal sites &
\footnotesize surgical spread \\
\end{tabular}
\caption{The UART-1 case study in real routed databases: the
$8\times8\,\mu$m defect window, its bond-pad layer, and the full die; top,
baseline and rejected reroute; bottom, the committed repair; red boxes mark
DRC violations.}
\label{fig:case}
\end{figure}

\paragraph{The picosecond attribution and a tight-clock flip.}
A separate attribution study replays committed edits through
\ecoroute{} while holding every untouched net byte-identical; a
step-by-step I2C walkthrough is in Appendix~L.  On a UART
resize, the replay reroutes only the resized instance's two nets, holds
that checkpoint's other 838 nets byte-identical, reaches DRC0, and moves max WNS by one
picosecond, where the historical full-route pass reported a $+0.795$\,ns
setup shift after rewiring the clock tree.
Re-judging both frozen results under a tightened clock exhibits a
$0.794$\,ns decision-flip band: any setup target inside it fails the
full-reroute result and passes the minimal-disturbance one.

\subsection{When Visual Evidence Helps}

In the development study of relocation-target choice (Appendix~M), the pixel-free choice lands on occupied geometry and is
displaced by legalization on both designs, while the \text{V-Pix}
occupancy instrument of Section~\ref{subsec:generate} lands exactly: 2/2
versus 0/2.  Both \text{V-Pix} repairs end at DRC\,6 and are rejected;
the supported signal is placement occupancy, never routing, timing, or
DRC.

Three preregistered held-out campaigns reuse the policy frozen before any
held-out pixel was rendered.  At five hash-random anchors per design,
\text{V-Pix} and pixel-free \text{N-Hash} chose different ring points
but both legalized every relocation (5/5 twice): random neighborhoods
saturate the binary feasibility gate.  At five DB-selected contested
anchors, the registered edge returned: \text{V-Pix} 4/5 versus
\text{N-Hash} 3/5, with all ten outcomes predicted by occupancy.
Thus pixels help at contested landings, not random anchors.

\subsection{Held-Out Diagnosis on GCD}

A clean GCD foundation receives five held-out seeds with 13 HBT-driver
overdrives plus verifier-only decoys.  Six methods run per seed (Appendix~N).
All 30 outputs pass physical checks, but only the frozen agent diagnostic
restores all 13 clean masters exactly with no wrong edit.  Blanket downsizing
finds the 13 positives but restores four; random selection and stock repair
make false edits; the register-only Pin-3D rule misses the combinational
defect.  Per-seed overhead recovery is 29.1--51.1\% (mean 38.9\%).
One failed execution was corrected before outcomes were observed, and an
independent audit verified all 30 runs.

\section{Conclusion}
In this paper, 
we presented \system{}, an agentic framework that closes the 3D
physical-ECO loop behind fail-closed gates: typed evidence drives
\text{Advance}, \text{Query}, or \text{Revise}; \ecoroute{} executes
attributable edits; the verifier commits or rolls back.  It repairs
what full reroute, stock repair, and the single agent leave standing,
at a fraction of their disturbance, and holds across two backbones.

\bibliography{references_v2}

\clearpage
\appendix
\raggedbottom
\setcounter{dbltopnumber}{1}

\noindent This technical appendix documents the evidence behind the main
text without introducing any separate headline claim.  Appendices~A--H
record the infrastructure: the pinned platform, the released code package,
the agent contract cards and sealed prompts, the tool budget, the ledger
vocabulary, a worked trajectory, the rendering instrument, and a glossary.
Appendices~I--N follow the main text's experimental order with the raw
measurements behind each summary; Appendices~O and~P state the evaluation
boundaries and the extended related-work context.  The main text remains
self-contained; the appendix exists so that every mechanism can be audited
and every reported experiment reproduced from the released artifacts, with
no reliance on model-authored success fields or external links.

\section{Platform and Pinned Versions}
\label{app:platform}

Every claim in this appendix ultimately rests on a fixed physical
substrate, so the record begins there: the main paper's setup states that
the platform is pinned end to end and that exact commits and host are
listed here.
All experiments run on the public, unmodified \engine{} repository at
commit \texttt{a73758a} with OpenROAD build \texttt{305d3ba}, on Ubuntu
22.04 (two Xeon Gold 6426Y, 251\,GiB RAM).  The studied designs are GCD,
the macro-free \texttt{I2cGpioExpander} (I2C), and a UART, all on the
ASAP7 predictive PDK.  Every router, timer, and DRC invocation uses these
pinned binaries.

\section{Artifact: Code and Environment Release}
\label{app:artifact}

A pinned platform is only auditable if the code that ran on it ships with
the paper.  The released repository\footnote{Released at \coderepo{} (engine pinned as a git
submodule; clone with \texttt{git clone --recurse-submodules}).} contains the complete agent
framework and integration surface described in the methodology section of
the main paper; Figure~\ref{fig:repotree} maps its layout, organized so
that each guarantee is inspectable in code at the indicated path.

Readers can exercise the full loop with no model, no API key, and no EDA
engine: \texttt{python3} \path{scripts/smoke_test.py} runs one offline episode end
to end (a candidate is rejected by the DRC gate, rolled back, and a guarded
alternative accepted), re-verifies the sealed ledger from disk, and exits
nonzero on any failure.  Fixture-backed runs are watermarked
\path{scientific_claim_eligible=false} in their own ledgers.  The code is
released under the BSD-3-Clause license.

\begin{figure}[t]
\begin{lstlisting}[style=repotree]
AgenticECO/
  agenticeco/          # runtime: orchestrator,
                       #  ledger, schema, verifier
  framework/           # role roster as data: one
                       #  contract card per agent
  adapters/taiwei/     # EcoRoute runner, ledgers,
                       #  fail-closed gates (clock
                       #  nets excluded in code)
  engine_cc/           # single-agent control arm
  episodes/            # sealed manifest + example
  third_party/TaiWei/  # engine pinned as a git
                       #  submodule; versions.lock
  docker/, Dockerfile  # one-step self-check image
  docs/REPRODUCING.md  # reproduction protocol
  scripts/smoke_test.py # offline full-loop check
  tests/               # runtime + visual suites
\end{lstlisting}
\vspace{-4pt}
\caption{Layout of the released code package.  The submodule pin is the
exact commit set of Appendix~\ref{app:platform}; every guarantee named in
the methodology section is inspectable at the indicated path.}
\label{fig:repotree}
\end{figure}

\section{Agent Construction: Contract Cards and Prompt Palette}
\label{app:prompts}

With the platform pinned (Appendix~A) and the code released (Appendix~B),
the remaining trust question is who is allowed to say what.  The
methodology section of the main paper answers it with a claim: every agent
operates under a declared contract, and no model statement is ever treated
as a measurement.  This section reproduces the governing text behind that
claim.
Every agent in \system{} is governed by a \emph{contract card}: a declaration
of what it may see, what it must return, and what it may never do, plus a
prompt card stating its protocol.  The cards summarized below are the
governing text shipped in the released repository
(\path{framework/prompts/}; the orchestrator's card keeps its historical
internal filename \texttt{operator.md}); the runtime schema enforces the same
boundaries mechanically, so a prompt cannot widen what the schema admits.

\subsection{Orchestrator}
\label{app:prompt-orchestrator}

\paragraph{Contract.}  The orchestrator performs evidence-gated repair-program
synthesis.  It sits at the root of the loop (query $\to$ generate $\to$
execute $\to$ verify) and delegates across four read-only specialist roles:
netlist, layout\_vpix, hb\_graph, and drc\_router\_logs.  It emits a repair
specification and then a deterministic repair program $g$;
it never emits coordinates, tool verdicts, or measurements.

\paragraph{Protocol.}  (1)~Delegate read-only analyses to the specialists;
every returned claim must cite its source artifact.  No specialist statement
is a measurement.  (2)~Aggregate findings into a repair specification:
targeted violations, movable instances, protected objects led by the clock
tree, and the fixed acceptance criteria inherited from $\Gamma$; intent and
constraints only, never coordinates.
(3)~Synthesize the repair program $g$: a short deterministic procedure that
turns evidence into a ranked candidate list.  Sampling stops at
$g$; from $g$ onward the round is replayable.  (4)~After a failed attempt,
decide explicitly: \text{Advance} to the next ranked candidate,
\text{Query} further evidence, or \text{Revise} after judging the
registered family refuted.  Record the decision and its evidence.
(5)~The procedure may be
revised after a failure; the registered acceptance gates may never be
revised.

\paragraph{Failure modes to avoid.}  Resampling a refuted candidate family
instead of excluding it; escalating to full reroute without written
justification; treating a plateaued DRC count as unreachable without probing
invariance.

\subsection{Read-Only Specialists}
\label{app:prompt-specialists}

\paragraph{netlist: source and netlist semantics.}  Sees HDL source windows,
netlist windows, and the violating-net list; returns typed evidence records,
each citing file/line; never sees pixels, DRC totals, or timing numbers it
did not query.  Its role is to explain what the violating nets are:
their drivers, sinks, hierarchy, and function, so the orchestrator knows which
instances are movable and which objects must be protected.  It returns
findings only, each citing the exact source artifact.

\paragraph{layout\_vpix: rendered-layout occupancy (V-Pix).}  Sees one
rendered layout window (pixels only); returns
\path{visual_specialist_result/v1} records carrying tier, bounding box,
repair side, confidence, and enumerated evidence codes; never sees candidate
IDs, instance or net names, or DRC and routing outcomes.  Its role is to
score how occupied the placement and routing look around a
location.  The scope is deliberately narrow: exact-feasibility of a landing
site, nothing downstream.  It cannot name candidates; a frozen spatial
mapper converts its geometric signal into candidate risks outside the model.

\paragraph{hb\_graph: hybrid-bond conflict graph.}  Sees the bond-terminal
geometry of cross-tier nets and the spacing rule; returns conflict-graph
records of violating pairs, separations, and connected components; never
sees pixels or timing.  Its role is to build the conflict structure
underlying the defect set: which bond terminals violate spacing against
which, and whether surviving violations are coupled (removing one
re-creates another).  Coupling is what shifts the orchestrator from single
relocations to relocation tuples.

\paragraph{drc\_router\_logs: reports to typed records.}  Sees the DRC
reports and router logs of the baseline and of failed rounds; returns
normalized violation records and the escalation/round structure; never sees
pixels or source code.  Its role is to turn raw tool text into typed
evidence: violation type, layer, bounding box, the nets implicated, and what
the router did in each round (reroute-set growth, new violations).  Failed
rounds are evidence, not noise.

\subsection{Non-Model Contracts: Executor and Verifier}
\label{app:prompt-contracts}

The remaining two cards govern programs, not models; they are stated here
because they complete the trust chain.

\paragraph{\ecoroute{} executor.}  Input: one registered edit (instance
$\to$ new placement, same tier).  Substrate: a copy of the design
and the unmodified pinned router.  The initial reroute set is exactly the
nets incident to the moved instances; every other net is frozen and checked
net-by-net by geometry hash.  Escalation is bounded ($k\le3$) and admits
implicated signal nets only; clock nets are never admitted.  The output is
$X'$, a realized-edit record, and the disturbance ledger.  That ledger is a
certificate, not an estimate: each untouched net's routed geometry must hash
identically to baseline, across both tiers of the folded stack and including
bond-layer vias.  A failure costs nothing but its evidence
(copy-on-write).

\paragraph{Independent verifier.}  Frozen and external.  It reads baseline
and candidate tool artifacts and the gates registered before
execution; it never reads the model text of any agent.  The gates are
routing completion, DRC, max/min timing, structural equivalence, and
preservation.  A pass commits $X'$; a failure rolls back and
the artifacts join the evidence set.  A failure can change the agent's
procedure; it can never change the standard.  When the objective is
unreachable within the edit space, the loop returns a \emph{verified
failure} backed by the same artifacts as a success.

\subsection{The Sealed Single-Agent Prompt}
\label{app:prompt-single}

The single-agent control arms (both backbones) run under one sealed
prompt template, frozen before the first single-agent run; \texttt{\{...\}}
fields are filled per cell and nothing else changes.  Tools are listed
alphabetically to avoid ordering hints.  The template is reproduced
verbatim in Figure~\ref{fig:sealedprompt}.

\begin{figure*}[t]
\begin{promptbox}[unbreakable, title={Sealed single-agent prompt (template, both backbones)}]
You are an autonomous EDA engineer.  Repair a post-route defect in a
folded-3D (ASAP7, two-tier, hybrid-bond) design.  You work alone: you may
not spawn sub-agents or delegate.

\textbf{Workspace}: \{run\_dir\} (your scratch; the episode checkpoint has
been copied here).  Original episode reference: \{episode\_dir\}
(READ-ONLY --- never write there).

\textbf{Design}: \{design\}, episode \{episode\}.  Baseline (from sealed
signoff): DRC violations = \{drc\_baseline\}; max WNS = \{wns\_baseline\}
ps; TNS = \{tns\_baseline\} ps; HBT vias = \{hbt\_baseline\}; total nets =
\{total\_nets\}.  Baseline reports are in \{run\_dir\}/baseline/.

\textbf{Goal}, in priority order:
1.~DRC $=$ 0 (detailed-route violations fully cleared);
2.~timing not degraded: final max WNS $\ge \min(0,\{wns\_baseline\})$ ps
and TNS not worse than baseline;
3.~disturb as little of the existing layout as possible: fewer nets with
changed routing geometry is better; avoid touching clock nets;
4.~preserve power and hybrid-bond count as close to baseline as possible.

\textbf{Tools} (invocation details in TOOL\_REGISTRY.md, provided in
\{run\_dir\}): listed alphabetically --- drc\_check, eco\_route,
full\_reroute, gui\_render, move\_cells, power\_report, sta\_report.  You
may also inspect any file in your workspace with ordinary shell commands.

\textbf{Budget (hard)}: at most 40 tool invocations total; at most 3
signoff attempts (a signoff $=$ drc\_check $+$ sta\_report $+$
power\_report on a candidate result).  When the budget is exhausted, stop
and report your best state.

\textbf{Report} (final message): the final checkpoint path, and for each
claim (DRC, WNS, TNS, disturbed nets, clock nets, HBT, power) the path of
the tool output file that proves it.  Unverified claims are worthless: any
number you state without a corresponding tool output will be treated as
fabricated.
\end{promptbox}
\caption{The sealed single-agent prompt template, verbatim.  It is frozen
before the first single-agent run and shared by both backbones;
\texttt{\{...\}} fields are filled per cell and nothing else changes.}
\label{fig:sealedprompt}
\end{figure*}

\section{Tool Registry and Budget Accounting}
\label{app:tools}

The contract cards of Appendix~C name seven tools; this section prices
them.  Every method runs each case once under an identical
$\le\!40$-call, $\le\!3$-signoff budget, so the definitions below fix
what a call and a signoff are and make the call and signoff counts
reported in the main paper unambiguous.
Every arm sees the same seven tool names, alphabetical in the prompt and
listed with their budget semantics in Table~\ref{tab:toolreg}.  Each
resolves to a frozen wrapper that creates a new
timestamped run directory, records the SHA-256 of its inputs, and emits
machine-readable outputs; no wrapper can overwrite a previous run.

\begin{table}[t]
\centering
\footnotesize
\setlength{\tabcolsep}{3.5pt}
\begin{tabular}{@{}p{1.28in}p{1.87in}@{}}
\toprule
Tool (budget) & Semantics \\
\midrule
\texttt{eco\_route} (1 call) & minimal-disturbance route of a registered edit; rounds $k\le3$ \\
\texttt{move\_cells} (1 call) & legalized relocation of a named instance set on a DB copy \\
\texttt{full\_reroute} (1 call) & engine full detailed reroute (the ``hammer'') \\
\texttt{gui\_render} (1 call) & rendered layout window for V-Pix \\
\texttt{drc\_check}, \texttt{sta\_report}, \texttt{power\_report} (1 signoff attempt) & three faces of one signoff invocation: one fresh DEF reload $=$ one of the $\le3$ attempts \\
disturbance (exempt) & ledger/gate bookkeeping; reads DEFs, emits the per-net geometry ledger; not agent-visible \\
\bottomrule
\end{tabular}
\caption{Registered tools and their budget semantics; the 40-call budget
counts wrapper invocations.}
\label{tab:toolreg}
\end{table}

Two consequences of this accounting are worth stating.  First, a signoff is
deliberately atomic: an arm cannot buy a cheap DRC-only glance, since
asking about DRC costs the same attempt that also reveals timing and power,
which is what makes ``spend a signoff to prove timing is free'' (the UART-1
pivot of the main text) a real decision.  Second, the exempt disturbance
tool removes any incentive to avoid measuring one's own footprint: the host
audits the disturbance ledger of every candidate, so an arm cannot save
calls by skipping it.

\section{Evidence-Ledger Event Taxonomy}
\label{app:ledger}

A budget only binds if spending is recorded, so contracts and prices are
completed by bookkeeping: every episode, accepted or refused, joins an
audit trail, and this section gives the event vocabulary of that
tamper-evident record.
Each episode's ledger is a hash chain: every event carries the SHA-256 of
its canonical-JSON payload and of its predecessor, blobs are
content-addressed, and the writer API has no overwrite path (files are
opened create-new), so a finished record cannot be edited.  Sealing writes
the chain head; verification replays the chain from disk.  The event
vocabulary is in Table~\ref{tab:ledgertax}.

\begin{table}[t]
\centering
\footnotesize
\setlength{\tabcolsep}{3.5pt}
\begin{tabular}{@{}p{1.28in}p{1.87in}@{}}
\toprule
Event class & Recorded content \\
\midrule
\texttt{observe}          & scoped query, artifact hashes, checkpoint binding \\
\texttt{fanout.diagnose}  & delegation contracts, specialist reports, provider provenance (incl.\ \texttt{real\_subagent\_api\_used}) \\
\texttt{algorithm.exec} & frozen ranking/mapping subprocess: input hash, command hash, output \\
\texttt{plan.action\_sel} & registered program, ranked candidates, gates (all pre-execution) \\
\texttt{execute.completed} & executor receipt: realized edit, run directory, exit status \\
\texttt{fanout.verify} (legacy name) & independent non-model verifier verdict, gate-by-gate record \\
\texttt{decide} / \texttt{reflect} & \text{Advance}/\text{Query}/\text{Revise} decision with cited failure artifacts; rollback record \\
\texttt{seal}             & chain head, blob and event counts \\
\bottomrule
\end{tabular}
\caption{Ledger event classes.  Every accepted and rejected
candidate leaves the same event shape, which is what makes a verified
failure as auditable as a success.}
\label{tab:ledgertax}
\end{table}

Two provenance flags are carried end to end and decide claim eligibility:
\path{evidence_class} (\path{real_tool} vs.\
\path{synthetic_smoke}) and \path{real_subagent_api_used}.  A run
whose EDA side is a fixture can exercise every code path yet is watermarked
\path{scientific_claim_eligible=false} in its own ledger; the
released smoke test demonstrates exactly this.

\section{Worked Trajectory: The UART-1 Ledger, Round by Round}
\label{app:trajectory}

Appendices~C--E define the contract, the price list, and the bookkeeping;
this section shows all three operating at once on a single episode.
Figure~6 of the main text summarizes UART-1;
Table~\ref{tab:uarttrace} is
its complete budgeted tool-call ledger, transcribed from the archived run
(annotations condensed).  The budget column shows cumulative
calls/signoffs after each round; the seven exempt disturbance-ledger
audits that checked every candidate are not rows in the table.

Read in order, the ledger shows each budget decision doing work.  R1
spends three calls refuting the cheapest family first: rerouting alone
frees the crowded bond vias only to watch them re-collide at the same
sites ($9\!\to\!6$ and $9\!\to\!12$), so the family is excluded rather
than resampled.  R2 escalates to move-plus-reroute and plateaus at
DRC\,5; the first signoff (call 8) is then spent deliberately, and its
return, WNS bit-identical to baseline, converts ``displacement is
probably safe'' into a proven-free budget that later rounds may spend.
R3 is the contract's escalation discipline in miniature: a written
justification, one test of the hammer, and a rejection on evidence (DRC
14, 508/907 nets, 9/10 clock nets) rather than on taste.  R4 spends the
proven-free timing on the revised program, eight anchors spread to a
$\ge\!2\,\mu$m pitch, and reaches DRC $9\!\to\!1$ at 17/907 nets with
zero clock touches; R5 probes the residual pair and finds it invariant
under every family tried, which is what licenses reporting a floor
instead of guessing one.  The second signoff (call 13) certifies the
committed state.  The third signoff attempt is never spent: once the
residual is proven invariant, another certificate could not change the
decision, and the loop stops twenty-seven calls under budget.

\begin{table}[t]
\centering
\footnotesize
\setlength{\tabcolsep}{3.5pt}
\begin{tabular}{@{}llp{1.86in}l@{}}
\toprule
Round & Calls & What happened & Budget \\
\midrule
R1 & 1--3 & render; \texttt{eco\_route} alone, two variants: DRC
$9\to6$, $9\to12$; freed bond vias re-collide, family refuted & 3/40, 0/3 \\
R2 & 4--7 & \texttt{move\_cells}+\texttt{eco\_route}, two programs:
plateau at DRC 5; conflict pinned to the bond-site grid & 7/40, 0/3 \\
R2s & 8 & signoff of best-so-far (DRC 5): WNS bit-identical to
baseline; timing proven free & 8/40, 1/3 \\
R3 & 9 & written escalation; \texttt{full\_reroute}: DRC 14, 508/907
nets, 9/10 clock nets; strictly worse, rejected & 9/40, 1/3 \\
R4 & 10--11 & revised program spends the free timing: 8 anchors at
$\ge2\,\mu$m pitch; DRC $9\to1$, 17/907 nets, 0 clock & 11/40, 1/3 \\
R5 & 12 & probe the residual pair: it is move- and reroute-invariant
under every repair family tried so far & 12/40, 1/3 \\
R4s & 13 & signoff of committed state: WNS $+18.166$\,ps ($=$ baseline),
TNS 0, HBT 74, power $+0.024\%$, EQY pass & 13/40, 2/3 \\
\bottomrule
\end{tabular}
\caption{UART-1 budgeted ledger (arm \system{}, sealed episode), in call
order: the R4 signoff (call 13) follows the R5 invariance probe.}
\label{tab:uarttrace}
\end{table}

\section{Rendered-Layout Instrument Details}
\label{app:render}

\looseness=-1
One instrument has so far gone unpinned: the camera.  The visual evidence
used in the main paper is only as trustworthy as the
instrument that produced it, so this section pins that instrument the same
way Appendix~A pins the engine.
All layout imagery in the paper, both the V-Pix inputs and every figure
panel, comes from one pinned pipeline: the engine GUI run headless
(\texttt{xvfb}), loading an archived database and saving a
window-registered PNG at fixed resolution.  Disturbance renders hide all
routing/cut layers except the lit changed-net overlay (signal nets in one
highlight group, clock nets in a second), so lit area is exactly the
disturbance ledger drawn to scale; defect-window renders keep all layers.
DRC markers are drawn afterwards by an exact micron$\to$pixel affine map
from the violation report's bounding boxes; composition scripts crop and
annotate but never redraw layout pixels.  The released repository ships the
snapshot bundler with input hashing
(\path{adapters/taiwei/vision/}), and the V-Pix result must parse under
the narrow visual schema (tier, bounding box, side, confidence, enumerated
evidence codes), so a vision model cannot smuggle a candidate name through
free text into the loop.

\section{Glossary of 3D-IC Terms}
\label{app:glossary}

The infrastructure appendices close with vocabulary.  The main paper and
the sections above lean on the following 3D-IC terms; they are collected
here for readers outside physical design, after which
Appendices~I--N turn from mechanism to measurement.

\begin{description}
\item[Face-to-face (F2F)] two dies bonded with their metal stacks facing
each other; cross-tier nets cross the bond interface directly.
\item[Hybrid bond / HBT] the bonded pad pair realizing one cross-tier
connection; in our ASAP7 setup an HBT cut is $0.032\,\mu$m wide under a
$1.568\,\mu$m spacing rule, an asymmetry that lets legal cell
placements produce illegal terminals.
\item[Folded (merged) representation] both tiers and the bond layer
represented in one database and one routing stack, so a single router sees
cross-tier geometry.
\item[Bond-level cut-spacing defect] two HBT cuts of different nets closer
than the rule; the defect class studied here.  It threatens
manufacturability, not connectivity.
\item[ECO] engineering change order: a localized post-route repair that
must preserve the surrounding verified implementation.
\item[Disturbance ledger] the per-net certificate that untouched routing is
byte-identical to baseline (canonical geometry hash per net, both tiers,
bond vias included).
\item[Signoff] the fresh-reload evaluation of a candidate: full-design DRC
recount, max/min STA, power, and HBT count; one reload $=$ one budgeted
attempt.
\item[Verified failure] a terminal state backed by the same audited
artifacts as a success, produced when the objective is shown unreachable
within the edit space.
\end{description}

\section{Leakage-Controlled Agent Evaluation}
\label{app:cleanroom}

The experimental appendices begin with the rules of admission.  The main
paper reports every arm as a single official run against a sealed
case; this section documents the harness that makes such a run admissible,
and every study in Appendices~J--N runs inside it.
Agent benchmarks inside a live engineering repository are unusually vulnerable
to contamination.  A model may discover an earlier repair, a comparison table,
or a failed arm from the worktree, local memory, shell history, or another
agent's scratch space.  It may then appear to diagnose a defect that it has
actually seen before.  We therefore separate the method-visible
namespace from a controller-only evidence namespace
(Table~\ref{tab:cleanroom_controls}).  This isolation is
evaluation infrastructure, not an additional reasoning module in
\system{}.

\begin{table*}[t]
\centering
\footnotesize
\setlength{\tabcolsep}{4.0pt}
\begin{tabular}{@{}p{1.08in}p{2.62in}p{2.78in}@{}}
\toprule
Control & Implementation & Threat addressed \\
\midrule
Sealed input
& Byte-identical ODB, DEF, SDC, netlist, configuration, and required baseline
reports are copied into a read-only capsule and bound by SHA-256.
& Input drift, stale checkpoints, and unequal physical starting points. \\
\addlinespace
Filesystem
& Project roots, Git metadata, paper tables, old sessions, local memories,
plugins, web access, and previous method outputs are absent.  Only sealed
inputs and frozen wrappers are readable; only a new arm directory is writable.
& Retrieval of a prior answer, outcome, coordinate, victim scope, or repair
trace. \\
\addlinespace
Agent isolation
& Each method starts in a fresh session.  Delegated specialists start with no
parent conversation, receive disjoint lane tasks, cannot read the peer lane,
cannot call EDA, and cannot spawn descendants.
& Cross-arm leakage, anchoring between specialists, and hidden extra search. \\
\addlinespace
Commitment
& The orchestrator hashes both terminal specialist reports, then commits one
action and one ordered victim scope before any route result is visible.  The
first terminal routing outcome is final.
& Selecting an action or a favorable intermediate round after seeing physical
results. \\
\addlinespace
EDA wrappers
& EDA access is restricted to frozen wrappers with create-new run directories,
fixed eight-thread execution, explicit call budgets, and machine-readable
receipts.
& Unlogged tool calls, silent retries, thread-count confounding, and overwritten
failures. \\
\addlinespace
Host authority
& A host-side evaluator reconstructs hashes and call order and reads fresh
full-design DRC and max/min STA outputs.  Model-authored pass/fail fields are
never treated as physical authority.
& Self-reported success, incremental/full-DRC mixing, unit mistakes, and missing
evidence. \\
\bottomrule
\end{tabular}
\caption{Clean-room controls used by the agent-evaluation harness.  These
controls determine whether a run is admissible; they do not improve the
candidate repair.}
\label{tab:cleanroom_controls}
\end{table*}

\paragraph{Experience control and claim boundary.}
Every arm receives the same seven frozen tool wrappers, including the
\texttt{eco\_route} executor.  The clean-room single-agent baseline receives
no repair experience.  An \system{} arm may receive only a predeclared,
leak-scanned eight-rule cross-case playbook containing generic rules such as
``diagnose before execution'' and ``accept only after full signoff.''  It
contains no same-episode instance, coordinate, victim-net list, action, route
result, or comparison outcome.  This distinction is recorded as a method
resource rather than hidden.  Consequently, a comparison between the
experience-free single agent and the complete \system{} stack is an end-to-end
systems comparison, not a causal estimate of delegation alone.  A
paper-eligible causal study must additionally equalize experience or use a
factorial ablation over agent topology and experience access.

\paragraph{Specialist context isolation.}
The orchestrator creates a hash-bound query handoff before delegation.
Specialists may read only their own task, the sealed episode, the declared
experience bundle, and the exact query files named by that handoff.  They
cannot inspect the orchestrator's draft, the peer report, or any EDA outcome.
Their only write is a create-new lane report.  The orchestrator can read the
reports only after both specialists terminate.  Thus fan-out supplies two
independent diagnoses rather than two agents collaboratively converging on a
previously exposed answer.

\paragraph{Case construction and reuse across studies.}
All evaluation cases are natural post-route hybrid-bond defects: they
are induced by sweeping legitimate flow knobs (seed, \hbt{} site density,
utilization) until the frozen backend itself produces a defective
signoff state, and never by editing the database to plant a violation.
Injection risks encoding its own repair, namely the inverse of the injecting
edit, whereas knob-induced defects are states the production flow can actually
reach, with no privileged answer.  Each case is frozen with its checkpoint,
baseline signoff, and content hashes before any method runs, and every
method arm executes exactly once against the sealed state under one budget.
Because the manifest is sealed first and each cell is a single official run,
the same sealed cases, or their declared subsets, can serve every matched study (the method comparison,
the component ablations, and the backbone-substitution arms) as controlled
comparisons in which the manipulated factor is the only difference; reusing
cases across arms is what makes the columns comparable, and cannot tune any
method because no arm is ever rerun against an observed outcome.  The
component ablations use the I2C/UART subset: the GCD baseline already
violates timing, so its zero-tolerance non-regression gate converts
sub-picosecond extraction noise into gate flips, which would contaminate a
causal readout.  Freshly generated held-out cases are reserved for a future
generalization claim; no such claim is made here.

\paragraph{Specialist roles versus instantiation width.}
The four specialist roles of the methodology section (netlist semantics,
rendered layout, hybrid-bond conflict graph, and violation-report parsing)
are a query decomposition, not a fixed agent count.  The orchestrator
decides per round how many concurrent child agents to instantiate and which
roles each carries, under a sealed campaign bound: the matched matrix
permits at most three parallel specialist agents (observed widths range from three
on first-round diagnosis down to one on late convergent rounds), the
delegation ablation restricts the same tool face to width one in series,
and the clean-room protocol of this appendix pins exactly two disjoint-lane
specialists as a leakage-control constraint rather than a framework
constant.  Actual per-round width is recorded in each run's ledger.

\paragraph{Audit and resource accounting.}
Before launch, the harness scans the complete method-visible tree, including
printable strings embedded in binary inputs.  A paper-eligible run additionally
scans model-visible text, tool arguments and outputs, reasoning events, and
inter-agent messages after termination.  Forbidden content, a missing receipt,
a nonregistered EDA call, an incorrect thread count, a budget overrun, or a
hash mismatch invalidates the protocol instead of becoming a favorable or
missing table entry.  Native per-session usage events are aggregated over the
orchestrator and every child into input, cached-input, output, reasoning, and
total tokens.  Structured event logs separately count platform calls and
registered physical calls; launcher and EDA timestamps provide controller and
physical wall time.

The isolation launcher, leak scanner, hash verifier, usage aggregator, and
final evaluator are measurement apparatus.  They are not agents in
\system{}, cannot propose or modify a repair, and their outputs are never
returned to a running method.  We report method-tree resources for the
orchestrator and children, while keeping host-side instrumentation separately
identifiable.

\paragraph{Status of the preliminary live pilot.}
The current UART two-row pilot was deliberately labeled
\path{Preliminary-Test-Not-Paper-Eligible}.  Its prelaunch scan covered 55
method-visible files and found no forbidden match; the live \system{} session
created exactly two medium-reasoning specialists and issued three registered,
eight-thread physical calls.  It then failed closed at routing.  Because this
pilot did not run the formal post-termination finalizer, it demonstrates the
execution and measurement plumbing but is not used as paper-level efficacy
evidence.  Formal tables include only runs that satisfy the full pre- and
post-run protocol.

\section{Backbone-Substitution Arms: Per-Cell Results}
\label{app:backbone}

The first claim audited under that harness is transfer: the main paper
reports that the sealed contract carries across backbones,
giving the GPT-5.6-backbone rerun of Table~1 of the main text in
aggregate.  The per-cell tuples are in Table~\ref{tab:backbone-cells},
every number re-derived from raw artifacts
(fresh DRC recount at the final DEF, per-net ledger recount, sealed-baseline
re-hash) by an independent verification pass.  Gates: g1 $=$ DRC0, g2 $=$
disturbance ledger produced with zero clock nets touched, g3 $=$ timing
non-regression.

Three cell-level patterns underlie the aggregate.  First, both arms
clear the same nine defects, but they pay differently for the same
outcome: \system{}'s disturbed-net count is strictly lower than the
single agent's in six of nine cells, equal in two (I2C-2 and UART-2),
and higher only on GCD-1 (5/591 versus 3/591), so the mean gap (0.44\%
versus 1.00\%) is consistent per-cell frugality, not one outlier.
Second, the only budget violation in the whole rerun is disclosed in the
table note: the single agent on GCD-2 overran the call budget with 16
unregistered engine invocations and moved one bond via ($70\!\to\!71$);
the cell is reported as-is rather than repaired, and every other cell in
both arms preserves the baseline \hbt{} count exactly, with zero clock
nets touched anywhere.  Third, the g3 timing crosses concentrate
entirely on GCD, in all three single-agent cells and two of three
\system{} cells; as in the Opus 4.8 matrix, these are sub-picosecond
$\Delta$-timing trips on GCD's already-negative baseline, a property of
the design's frozen starting point rather than of either backbone.

\begin{table}[t]
\centering
\footnotesize
\setlength{\tabcolsep}{2.9pt}
\begin{tabular}{@{}ll rr r ccc r@{}}
\toprule
Arm & Cell & DRC in$\to$out & $\Delta$Nets & Clk & g1 & g2 & g3 & HBT \\
\midrule
Single & GCD-1  & 4$\to$0  & 3/591   & 0 & \checkmark & \checkmark & $\times$ & 74 \\
Single & GCD-2$^{a}$  & 6$\to$0  & 10/619  & 0 & \checkmark & \checkmark & $\times$ & 71 \\
Single & GCD-3  & 26$\to$0 & 16/614  & 0 & \checkmark & \checkmark & $\times$ & 75 \\
Single & I2C-1  & 5$\to$0  & 6/1247  & 0 & \checkmark & \checkmark & \checkmark & 97 \\
Single & I2C-2  & 23$\to$0 & 12/1245 & 0 & \checkmark & \checkmark & \checkmark & 80 \\
Single & I2C-3  & 6$\to$0  & 5/1248  & 0 & \checkmark & \checkmark & \checkmark & 84 \\
Single & UART-1 & 9$\to$0  & 8/907   & 0 & \checkmark & \checkmark & \checkmark & 74 \\
Single & UART-2 & 5$\to$0  & 4/882   & 0 & \checkmark & \checkmark & \checkmark & 62 \\
Single & UART-3 & 10$\to$0 & 10/944  & 0 & \checkmark & \checkmark & \checkmark & 60 \\
\midrule
Ours & GCD-1  & 4$\to$0  & 5/591   & 0 & \checkmark & \checkmark & $\times$ & 74 \\
Ours & GCD-2  & 6$\to$0  & 1/619   & 0 & \checkmark & \checkmark & $\times$ & 70 \\
Ours & GCD-3  & 26$\to$0 & 2/614   & 0 & \checkmark & \checkmark & \checkmark & 75 \\
Ours & I2C-1  & 5$\to$0  & 3/1247  & 0 & \checkmark & \checkmark & \checkmark & 97 \\
Ours & I2C-2  & 23$\to$0 & 12/1245 & 0 & \checkmark & \checkmark & \checkmark & 80 \\
Ours & I2C-3  & 6$\to$0  & 1/1248  & 0 & \checkmark & \checkmark & \checkmark & 84 \\
Ours & UART-1 & 9$\to$0  & 6/907   & 0 & \checkmark & \checkmark & \checkmark & 74 \\
Ours & UART-2 & 5$\to$0  & 4/882   & 0 & \checkmark & \checkmark & \checkmark & 62 \\
Ours & UART-3 & 10$\to$0 & 2/944   & 0 & \checkmark & \checkmark & \checkmark & 60 \\
\bottomrule
\multicolumn{9}{@{}l}{\scriptsize $^{a}$overran the call budget (16 unregistered engine}\\
\multicolumn{9}{@{}l}{\scriptsize \hphantom{$^{a}$}invocations) and moved one bond via (70$\to$71), reported}\\
\multicolumn{9}{@{}l}{\scriptsize \hphantom{$^{a}$}as-is; every other cell preserves the baseline HBT count.}
\end{tabular}
\caption{GPT-5.6-backbone per-cell results: Single $=$ the
single-agent control, Ours $=$ \system{}, identical sealed contract and
budgets.}
\label{tab:backbone-cells}
\end{table}

\section{Matched Component-Ablation Data}

\begin{figure}[t]
\centering
\includegraphics[width=\columnwidth]{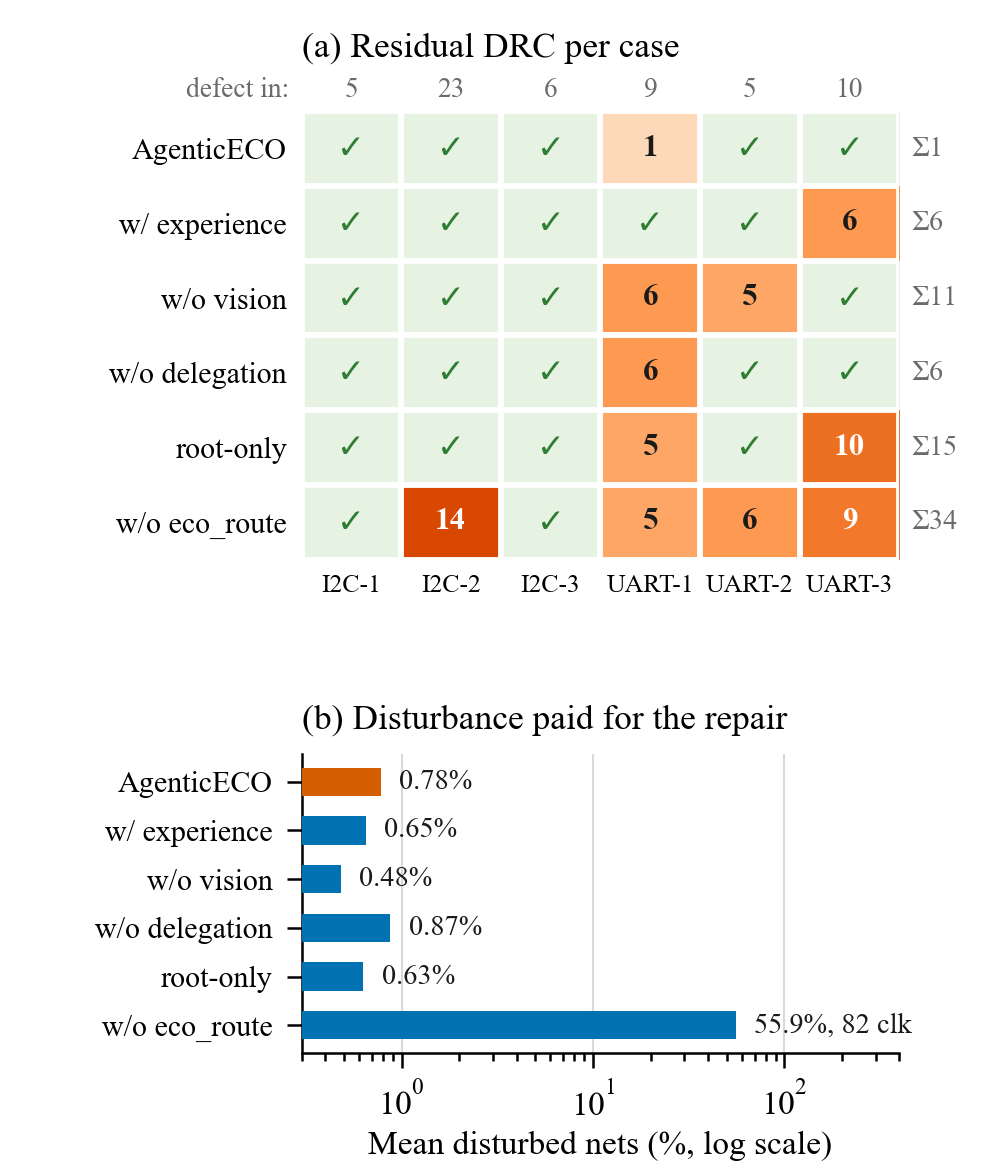}
\caption{Matched six-case component ablation, visualizing
Table~\ref{tab:ablation-raw}.  (a) Residual DRC per case and variant;
green marks a cleared case.  (b) Mean disturbed nets, log scale.}
\label{fig:ablation}
\end{figure}

\begin{figure}[t]
\centering
\includegraphics[width=\columnwidth]{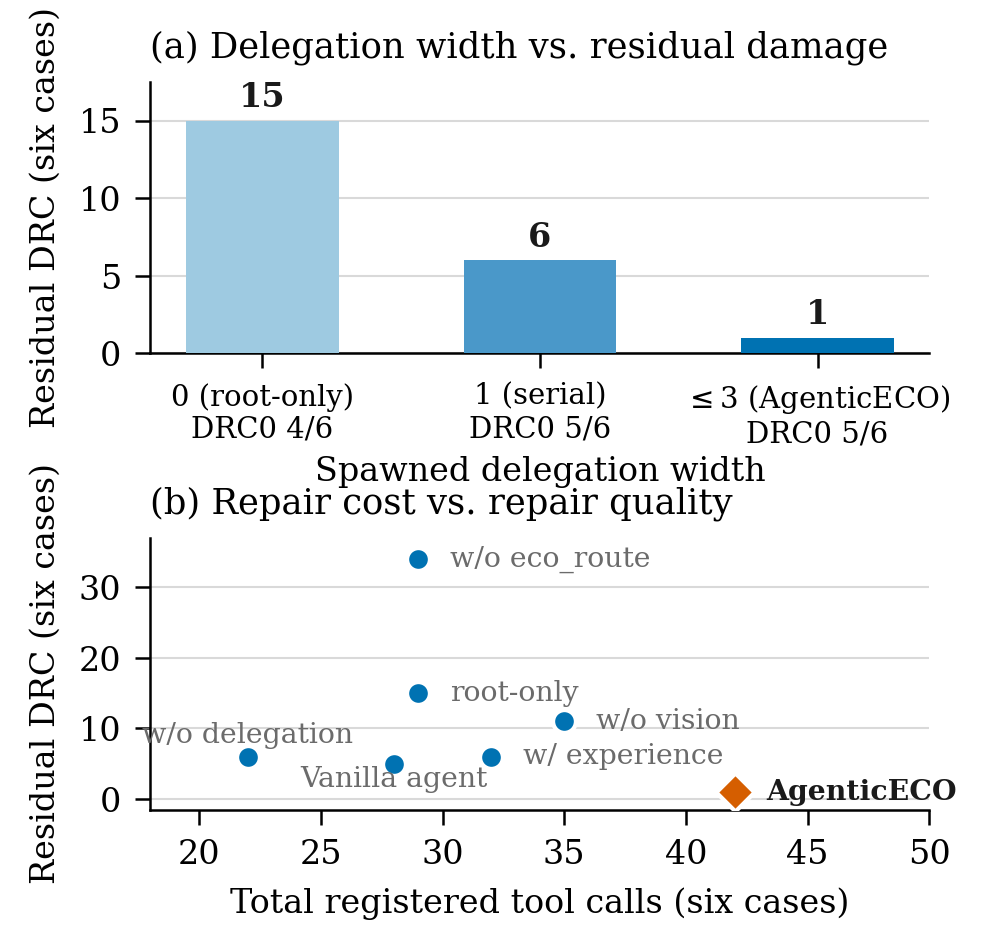}
\caption{Campaign statistics over the six cases: (a) residual DRC falls
as delegation width grows; (b) the full system spends most, damages
least.}
\label{fig:mm2stats}
\end{figure}

Where Appendix~J holds the architecture fixed and swaps the backbone,
this section does the opposite, removing one component at a time.  The
main paper's ablation paragraphs report which component removal costs
which case; this section is the raw evidence behind them.
Table~\ref{tab:ablation-raw} is the per-case table,
Figure~\ref{fig:ablation} visualizes it, and Figure~\ref{fig:mm2stats} adds
the cost--quality view over the same six-case configurations.

The raw columns support three readings that the summary bars compress.
First, cost and quality move together: the full system spends the most
registered tool calls (42 over the six cases) and leaves the least
damage ($\Sigma$DRC $58\!\to\!1$), while the serial w/o-delegation arm
is the cheapest (22 calls) and fails exactly the hardest site (UART-1 at
DRC\,6); calls are being spent where defects are contested, not evenly.
Second, the executor ablation is categorical rather than marginal: its
per-case disturbances (501--707 nets) sit two orders of magnitude above
every other arm's, and it is the only variant that touches clock nets
(82, against zero in all five others), which is what ``collapses onto
the hammer'' means at column level.  Third, the vision ablation posts
the lowest mean disturbance of all (0.48\%) while clearing only 4/6:
the two contested UART sites it fails (residuals 6 and 5) are precisely
where deliberate repair work, and hence disturbance, would have been
spent.  The experience arm matches the full system's 5/6 at lower cost
(32 calls, one signoff per case) but relocates the hard case, uniquely
zeroing UART-1 while leaving DRC\,6 on UART-3.

\begin{table*}[t]
\centering
\setlength{\tabcolsep}{1.35pt}
\scriptsize
\begin{threeparttable}
\begin{tabular}{l cc rrr rrr rrrrrrrrr}
\toprule
& \multicolumn{2}{c}{\textbf{Configuration}}
& \multicolumn{3}{c}{\textbf{I2C (three cases)}}
& \multicolumn{3}{c}{\textbf{UART (three cases)}}
& \multicolumn{9}{c}{\textbf{All six cases}} \\
\cmidrule(lr){2-3}\cmidrule(lr){4-6}\cmidrule(lr){7-9}
\cmidrule(lr){10-18}
\textbf{Variant}
& \textbf{Vision}
& \shortstack{\texttt{eco\_}\\\texttt{route}}
& \shortstack{Full\\DRC out\\(1/2/3)}$\downarrow$
& \shortstack{$\Delta$Nets\\(1/2/3)}$\downarrow$
& \shortstack{DRC0\\cases}$\uparrow$
& \shortstack{Full\\DRC out\\(1/2/3)}$\downarrow$
& \shortstack{$\Delta$Nets\\(1/2/3)}$\downarrow$
& \shortstack{DRC0\\cases}$\uparrow$
& \shortstack{$\Sigma$DRC\\in\\$\rightarrow$out}$\downarrow$
& \shortstack{DRC\\reduc-\\tion}$\uparrow$
& \shortstack{Hard\\UART-1\\resid.}$\downarrow$
& \shortstack{DRC0\\cases}$\uparrow$
& \shortstack{All\\gates\\pass}$\uparrow$
& \shortstack{Mean\\$\Delta$Nets\\(\%)}$\downarrow$
& \shortstack{Clock\\touch-\\ed}$\downarrow$
& \shortstack{Tool\\calls}$\downarrow$
& \shortstack{Sign-\\offs\\/case}$\downarrow$ \\
\midrule
\textbf{AgenticECO}
& \checkmark & \checkmark
& \textbf{0/0/0} & 3/12/4 & \textbf{3/3}
& \textbf{1/0/0} & 17/2/10 & \textbf{2/3}
& \textbf{58$\rightarrow$1} & \textbf{98.3\%} & 1
& \textbf{5/6} & \textbf{5/6} & 0.78 & \textbf{0} & 42 & 1.17 \\
w/ experience$^{*}$
& \checkmark & \checkmark
& \textbf{0/0/0} & 5/12/7 & \textbf{3/3}
& 0/0/6 & 10/4/4 & \textbf{2/3}
& 58$\rightarrow$6 & 89.7\% & \textbf{0}
& \textbf{5/6} & \textbf{5/6} & 0.65 & \textbf{0} & 32 & \textbf{1.00} \\
w/o \texttt{eco\_route}
& \checkmark & --
& 0/14/0 & 707/678/699 & 2/3
& 5/6/9 & 515/501/515 & 0/3
& 58$\rightarrow$34 & 41.4\% & 5
& 2/6 & 2/6 & 55.88 & 82 & 29 & 1.50 \\
w/o vision
& -- & \checkmark
& 0/0/0 & 3/12/3 & 3/3
& 6/5/0 & 8/1/4 & 1/3
& 58$\rightarrow$11 & 81.0\% & 6
& 4/6 & 4/6 & \textbf{0.48} & \textbf{0} & 35 & \textbf{1.00} \\
w/o delegation
& \checkmark & \checkmark
& 0/0/0 & 5/12/5 & 3/3
& 6/0/0 & 8/6/18 & 2/3
& 58$\rightarrow$6 & 89.7\% & 6
& \textbf{5/6} & \textbf{5/6} & 0.87 & \textbf{0}
& \textbf{22} & 1.33 \\
root-only$^{*}$
& \checkmark & \checkmark
& 0/0/0 & 5/12/5 & 3/3
& 5/0/10 & 8/7/3 & 1/3
& 58$\rightarrow$15 & 74.1\% & 5
& 4/6 & 4/6 & 0.63 & \textbf{0} & 29 & \textbf{1.00} \\
\bottomrule
\end{tabular}
\begin{tablenotes}[flushleft]
\scriptsize
\item[*] ``w/ experience'' and ``root-only'' are independently verified
re-run arms: same sealed cases, budgets, and Claude Opus 4.8 medium backbone as
the matrix.  Transcript-audited delegation widths: ``w/ experience'' fanned
out three concurrent specialists on two of six cases and worked alone
otherwise; ``w/o delegation'' runs one serial specialist agent; ``root-only'' spawns
no child agents by construction.
\end{tablenotes}
\end{threeparttable}
\caption{Matched architecture ablation on the six I2C/UART cases (raw data
for the main-paper ablation figure).}
\label{tab:ablation-raw}
\end{table*}

\section{Surgical Repair Attribution on I2C}

\begin{figure*}[!t]
\centering
\includegraphics[width=.78\textwidth]{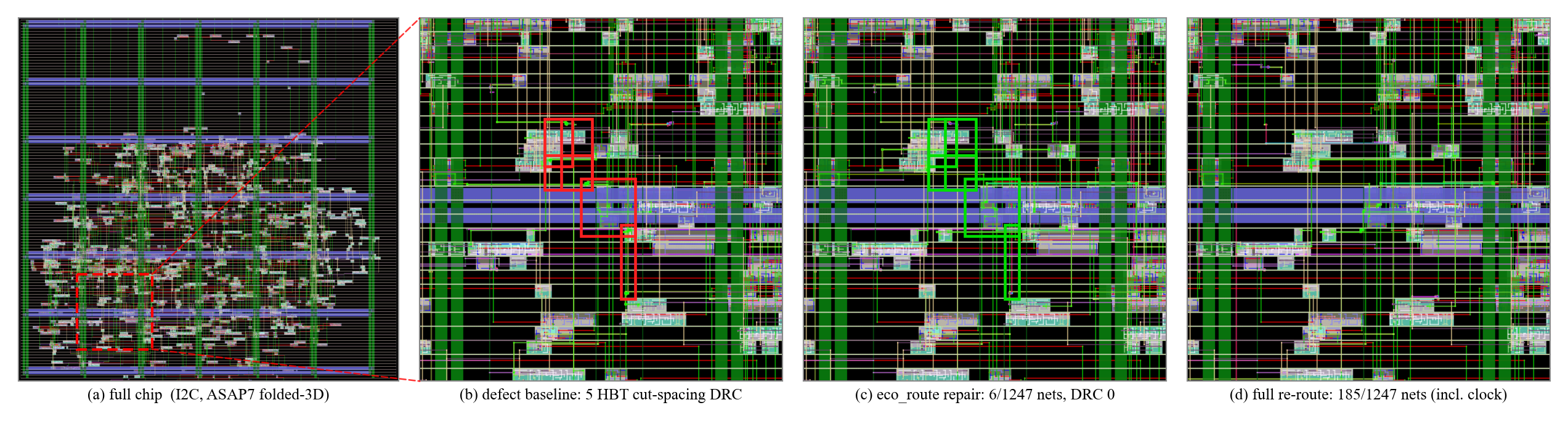}
\caption{Surgical repair versus full reroute on the natural I2C defect.
(a) Full-die routed baseline.  (b) Zoom on the five HBT cut-spacing
violations.  (c) After \ecoroute{}: only the six victim nets of the
three relocated split buffers change; the clock tree is untouched.
(d) After the no-move full reroute: churn spreads across the fabric and
rewires the clock tree.}
\label{fig:reveal}
\end{figure*}

The ablations above price whole components in disturbed nets; the next two
sections descend to a single repair and ask what its individual moves cost
physically.  The main paper attributes a picosecond-level timing effect to
a UART resize
through an \ecoroute{} replay; the companion I2C study below
decomposes the same mechanism on a natural three-relocation repair,
through \ecoroute{} replays that hold every untouched net
byte-identical.  Replaying the accepted
three relocations and rerouting only their six victim nets reaches DRC0 while
leaving the other 1,241 nets, the entire clock tree included, byte-identical
(Figure~\ref{fig:reveal}c), whereas the no-move full reroute escapes only
through the 56.7\% rewrite and bond re-plan of Figure~\ref{fig:reveal}d.  A
no-move control rerouting only the five violating nets stalls at one residual
violation between two of the split buffers the accepted repair relocates,
showing that relocation is necessary
in the minimal-disturbance regime.  A matched nonvisual control of ten
preregistered direction tuples rejected eight at the exact-placement check
(the M0 gate: a relocation must land on its chosen legal site with zero
legalization shift) and
saw its two legal hits each rewire a clock net, so occupancy-aware choice buys
legal, minimally disruptive landings, not the possibility of repair.
Table~\ref{tab:crossdesign} collects these cross-design outcomes.

\begin{table*}[t]
\centering
\footnotesize
\setlength{\tabcolsep}{4.0pt}
\begin{tabular}{@{}llp{1.72in}rrrrl@{}}
\toprule
Evidence & Design/arm & Edit or task receipt & DRC & \shortstack{Max/min\\WNS (ns)} & EQY & \shortstack{Power\\($\mu$W)} & Decision \\
\midrule
Pixel & I2C V-Pix & 1 relocation; exact, shift 0.000 $\mu$m & 6 & +15.612/+0.029 & PASS & 44.277 & Reject \\
Pixel control & I2C N-Hash & 1 relocation; shifted 0.162 $\mu$m (3 sites) & -- & -- & -- & -- & Reject M0 \\
Pixel & UART V-Pix & 1 relocation; exact, shift 0.000 $\mu$m & 6 & +11.385/+13.376 & PASS & 1390.410 & Reject \\
Pixel control & UART N-Hash & 1 relocation; shifted 0.216 $\mu$m (4 sites) & -- & -- & -- & -- & Reject M0 \\
\addlinespace
Natural & I2C TaiWei & 0 edits; defect diagnosed after full DRT & 5 & +15.614/-- & -- & 44.3$^*$ & Reject \\
Reflection & I2C graph+independent & 3 relocations; residual triangle remains & 3 & +15.610/-- & PASS & 44.286 & Reject \\
Reflection & I2C graph+joint+reroute & 3 exact E/W/E relocations; fresh GRT/DRT & 0 & +15.611/+0.030 & PASS & 44.278 & Accept \\
\addlinespace
Supporting & UART guarded resize & 1 same-tier \texttt{BUFx3}$\rightarrow$\texttt{BUFx2} & 0 & +12.697/-- & PASS & 1376.935 & Accept$^{\dagger}$ \\
\bottomrule
\end{tabular}
\caption{Cross-design physical outcomes; dashes mean not run or
unavailable.  N-Hash stopped after failing the exact-placement (M0) gate.
Power is absolute; $^*$the native I2C report
is rounded.  $^{\dagger}$The UART resize is setup/max-only supporting evidence;
its $-0.02131\%$ outcome is measured against an independently recovered
baseline, not a single-factor causal gain.}
\label{tab:crossdesign}
\end{table*}

\subsection{Full Walkthrough: Replaying the Accepted I2C Repair}
\label{app:i2c-walkthrough}

The accepted I2C repair is three same-tier relocations of split-buffer
instances that clear the natural five-violation defect.  The attribution
question is what those three moves cost by themselves, separated
from router churn.  The replay answers it in four steps, each leaving a
hashed artifact.

\paragraph{Step 1: reconstruct a routed starting state.}
The historical repair's own candidate database is unusable for
attribution: its flow tore up all routing before rerouting (1{,}241 of
1{,}247 nets unrouted), so replaying it would measure a full reroute, not
the edit.  Instead the replay starts from the routed baseline
checkpoint and applies exactly the three recorded relocations
(coordinates and orientations bit-identical to the accepted repair) with
routing untouched.

\paragraph{Step 2: a fail-closed rejection that proves the gate.}
The first replay attempt fed the historical candidate database anyway,
and the per-round gate correctly refused it: from the frozen baseline's
perspective, 1{,}239 nets of routing would have ``appeared from nowhere''
(1{,}239 is the 1{,}247-net total minus the eight nets that are unrouted
even in the baseline, whose absence the gate does not count; the candidate
database itself keeps routing on only six nets, hence its 1{,}241
unrouted).
This rejected run is kept as evidence that the freeze contract fails
closed rather than silently absorbing a full reroute.  (Incidentally, that
de-facto full reroute converged to three residual violations where the
same input historically reached zero; full reroute is sensitive to
environment and iteration order, one more reason its outcomes cannot be
attributed to an edit.)

\paragraph{Step 3: the minimal-disturbance replay.}
With the correct starting state, \ecoroute{} reroutes only the six nets
incident to the three moved instances and freezes the other 1{,}241,
verified net-by-net from output geometry
(Table~\ref{tab:i2c-replay}):

\begin{table}[t]
\centering
\footnotesize
\setlength{\tabcolsep}{3.5pt}
\begin{tabular}{@{}lrr@{}}
\toprule
 & Full reroute & \ecoroute{} replay \\
\midrule
Nets changed            & 185/1247 & \textbf{6/1247} \\
Clock nets rewired      & 11 (incl.\ root) & \textbf{0} \\
Natural DRC             & $5\to0$ & $5\to0$ \\
New DRC                 & 0 & 0 \\
Setup WNS (ps)          & $+15611.2$ & $\mathbf{+15613.7}$ \textbf{(=base)} \\
Hold WNS / TNS / FEP    & $+29.82$ / 0 / 0 & $+27.96$ / 0 / 0 \\
HBT physical vias       & 93 ($-4$) & \textbf{97 (=base)} \\
Total power ($\mu$W)    & 44.27829 & 44.27677 \\
Detailed-route wall (s) & 103.8 & 49.7 \\
\bottomrule
\end{tabular}
\caption{Same checkpoint, SDC, and host: three relocations judged through
two executors.  The replay's WNS is the baseline value because its
1{,}241 frozen nets keep their timing.}
\label{tab:i2c-replay}
\end{table}

Both executors clear the defect; only one produces a result attributable
to the edit.  Every violation implicated at least one rerouted net, so the
clearance is visible to the checker as a true elimination, not frozen out
of view.

\paragraph{Step 4: what the replay taught the gate.}
The I2C scale exposed one over-strict rule in the first gate version: the
baseline itself contains eight dangling nets with no routing, and the gate
demanded they appear as frozen routing.  The corrected rule was added with
two regression tests: an unrouted baseline net is preserved by staying
unrouted, while growing routing from nothing remains a failure.
The episode is recorded because it is the honest shape of a freeze
contract meeting a real design: the contract, not the measurement, is what
gets amended, and only ever in the stricter-to-looser direction visible in
the released \texttt{eco\_gate.py}.

\paragraph{Scope.}
On a design this small the wall-clock advantage ($2.1\times$) is
incidental, since detailed-route fixed costs dominate.  The walkthrough's
claim is the disturbance and timing-preservation contrast, and it is the
mechanism behind the picosecond attribution of the main text.

\section{Visual Feasibility Campaigns}

\begin{figure}[t]
\centering
\includegraphics[width=\columnwidth]{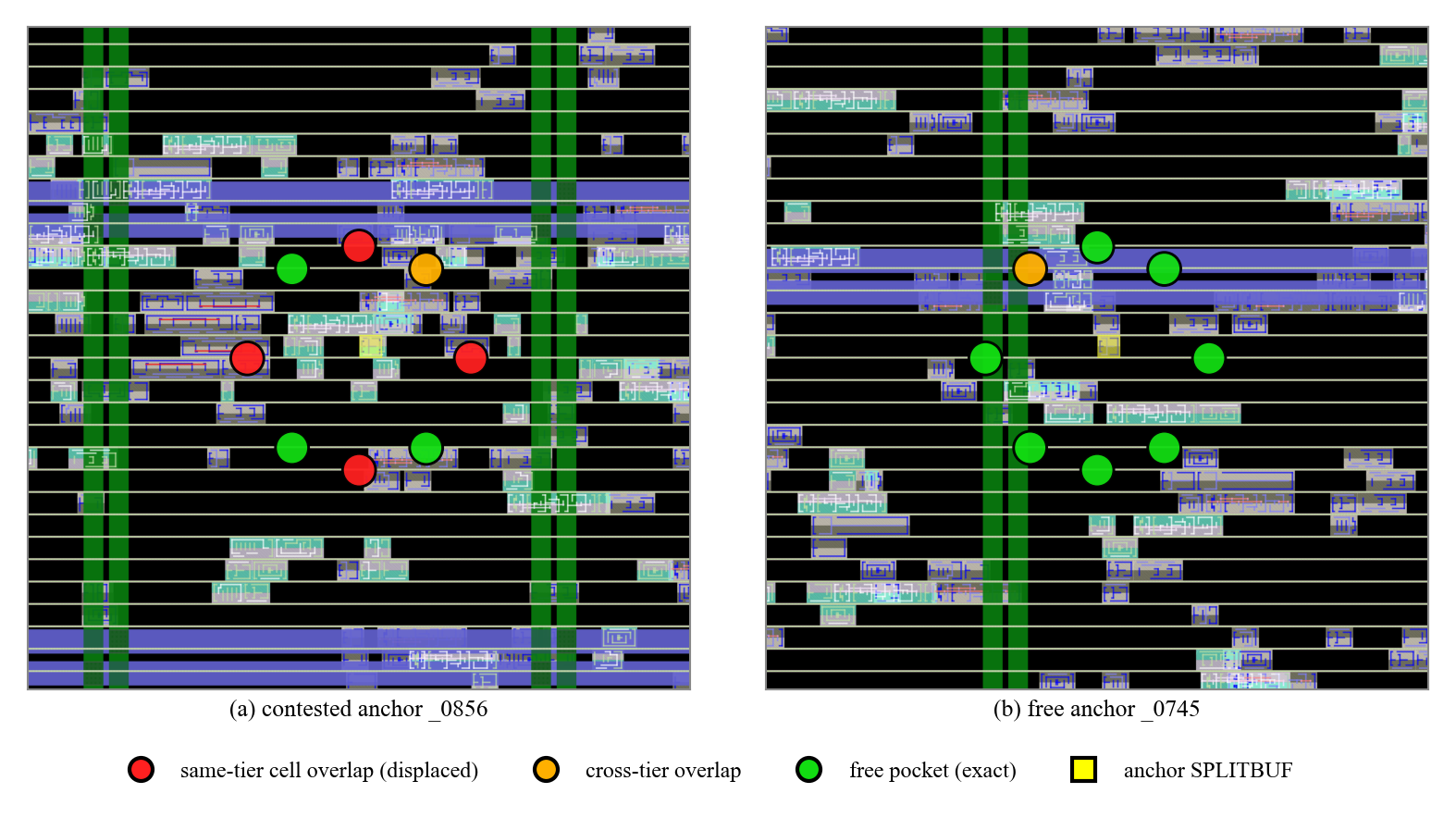}
\caption{Contested versus hash-random anchors.  Where ring landing points
overlap cells, pixels regain an exact-feasibility edge; where the ring is
clean, the gate saturates.}
\label{fig:anchors}
\end{figure}

The repairs replayed above were selected with visual evidence in the loop;
this section collects every campaign in which that evidence was itself put
on trial.  The ``When Visual Evidence Helps'' subsection of the main text
summarizes
these campaigns; Table~\ref{tab:heldout_campaigns} lists every one of them
and Figure~\ref{fig:anchors} shows the anchor geometry that decides the
outcome.  Three reading notes complete that summary.  First, the
development study is a placement-occupancy result only: both full visual
candidates end at DRC 6 and are rejected, so the supported signal is never
routing, timing, or DRC.  Second, the held-out UART row carries one
development-exposed anchor, disclosed in the table note, so that campaign's
held-out claim rests on 4/4 versus 4/4 rather than on 5/5.  Third, the
contested campaign is the registered prediction rather than a post-hoc
slice: the five contested anchors were chosen by a pixel-free DB-geometry
census, and all ten of its outcomes are predicted exactly by occupancy of
the chosen point.  Read against Figure~\ref{fig:anchors}, the scope is the
registered one: pixels confer an exact-feasibility edge only where
landings are contested, and hash-random anchors are not contested.

\begin{table}[t]
\centering
\footnotesize
\setlength{\tabcolsep}{3.5pt}
\begin{tabular}{@{}llcc@{}}
\toprule
Campaign & Anchors & V-Pix & N-Hash \\
\midrule
Development (2 designs) & defect-site & 2/2 & 0/2 \\
Held-out I2C            & hash-random & 5/5 & 5/5 \\
Held-out UART           & hash-random & 5/5$^{a}$ & 5/5$^{a}$ \\
Held-out I2C, registered pred. & contested & \textbf{4/5} & 3/5 \\
\bottomrule
\multicolumn{4}{@{}l}{\scriptsize $^{a}$one anchor development-exposed and disclosed;}\\
\multicolumn{4}{@{}l}{\scriptsize \hphantom{$^{a}$}held-out claim rests on 4/4 vs 4/4.}
\end{tabular}
\caption{Visual-feasibility campaigns: exact-relocation gate, one
relocation per arm, anchors and choices sealed before EDA.  Pixels
discriminate only at occupied landings.}
\label{tab:heldout_campaigns}
\end{table}

\section{Held-Out GCD Diagnosis Details}

Every study so far lives on the I2C/UART development matrix; the held-out
GCD diagnosis leaves it.
The main paper's held-out GCD subsection reports method-level aggregates and
points here for the per-method and per-seed detail.  All 30 physical results pass
the pinned OpenROAD and structural-equivalence
checks.  The Agent diagnostic is the only method that restores all 13 clean
masters exactly without a wrong instance or target.  Recovery remains below
100\% because the denominator includes verifier-only non-HBT power decoys
outside the frozen diagnostic policy.  The register-only Pin-3D rule scans
launch registers by construction (its discovery logs enumerate them per
seed) and finds none actionable, so it emits no edit anywhere (0/0/13):
every injected overdrive sits on a combinational \hbt{}-driver, which is
the main text's ``misses the combinational defect.''

\begin{table*}[t]
\centering
\footnotesize
\setlength{\tabcolsep}{4.2pt}
\begin{tabular}{lrrrrr}
\toprule
Method & Physical pass & TP/FP/FN & Exact & Wrong depth/master & Mean recovery \\
\midrule
Agent diagnostic & 5/5 & 13/0/0 & 13/13 & 0 & +38.87\% \\
Blanket one-step down & 5/5 & 13/0/0 & 4/13 & 9 & +22.37\% \\
Random matched-count & 5/5 & 2/11/11 & 0/13 & 2 & $-19.19\%$ \\
Stock timing repair & 5/5 & 0/20/13 & 0/13 & 0 & $-19.06\%$ \\
Pin-3D rule & 5/5 & 0/0/13 & 0/13 & 0 & 0.00\% \\
No repair & 5/5 & 0/0/13 & 0/13 & 0 & 0.00\% \\
\bottomrule
\end{tabular}
\caption{Five held-out GCD seeds, ten HBT candidates each, 13 positives.
TP/FP/FN scores identification; Exact and Wrong depth/master score the
realized cell. Recovery is $(P_0-P_m)/(P_0-P_{\mathrm{clean}})$ over
no-repair and method powers: injected-overhead recovery, not a 38.87\%
chip-power reduction; the Agent's mean total-power drop is 0.09336\%
(566.76 nW).}
\label{tab:gcd-diagnosis}
\end{table*}

\subsection{The Full 30-Cell Grid: Five Seeds $\times$ Six Methods}
\label{app:gcd-grid}

Table~\ref{tab:gcd-diagnosis} aggregates over seeds; the complete grid is
Table~\ref{tab:gcd-grid}.  Each cell reports corrupted-instance identification (TP/FP/FN
against the hidden ground truth), exact master restorations, and
injected-overhead recovery on total power.  Positive counts differ by seed
(3/2/2/2/4); recovery $0.00\%$ marks the two arms that emit no edit.

\begin{table*}[!t]
\centering
\footnotesize
\setlength{\tabcolsep}{4.2pt}
\begin{tabular}{@{}l rrr rrr rrr rrr rrr@{}}
\toprule
& \multicolumn{3}{c}{\textbf{Seed 1} (3+)} & \multicolumn{3}{c}{\textbf{Seed 2} (2+)}
& \multicolumn{3}{c}{\textbf{Seed 3} (2+)} & \multicolumn{3}{c}{\textbf{Seed 4} (2+)}
& \multicolumn{3}{c}{\textbf{Seed 5} (4+)} \\
\cmidrule(lr){2-4}\cmidrule(lr){5-7}\cmidrule(lr){8-10}\cmidrule(lr){11-13}\cmidrule(lr){14-16}
Method & T/F/N & Ex & Rec.\% & T/F/N & Ex & Rec.\% & T/F/N & Ex & Rec.\%
       & T/F/N & Ex & Rec.\% & T/F/N & Ex & Rec.\% \\
\midrule
\textbf{Agent diagnostic}
& \textbf{3/0/0} & \textbf{3} & $\mathbf{+33.5}$
& \textbf{2/0/0} & \textbf{2} & $\mathbf{+35.3}$
& \textbf{2/0/0} & \textbf{2} & $\mathbf{+45.4}$
& \textbf{2/0/0} & \textbf{2} & $\mathbf{+29.1}$
& \textbf{4/0/0} & \textbf{4} & $\mathbf{+51.1}$ \\
Blanket one-step down
& 3/0/0 & 1 & $+20.5$
& 2/0/0 & 1 & $+21.9$
& 2/0/0 & 0 & $+19.6$
& 2/0/0 & 1 & $+18.0$
& 4/0/0 & 1 & $+31.9$ \\
Random matched-count
& 1/2/2 & 0 & $-8.2$
& 0/2/2 & 0 & $-26.5$
& 0/2/2 & 0 & $-29.4$
& 0/2/2 & 0 & $-23.2$
& 1/3/3 & 0 & $-8.7$ \\
Stock timing repair
& 0/4/3 & 0 & $-11.5$
& 0/4/2 & 0 & $-25.4$
& 0/4/2 & 0 & $-23.6$
& 0/4/2 & 0 & $-18.0$
& 0/4/4 & 0 & $-16.7$ \\
Pin-3D rule
& 0/0/3 & 0 & $0.0$
& 0/0/2 & 0 & $0.0$
& 0/0/2 & 0 & $0.0$
& 0/0/2 & 0 & $0.0$
& 0/0/4 & 0 & $0.0$ \\
No repair
& 0/0/3 & 0 & $0.0$
& 0/0/2 & 0 & $0.0$
& 0/0/2 & 0 & $0.0$
& 0/0/2 & 0 & $0.0$
& 0/0/4 & 0 & $0.0$ \\
\bottomrule
\end{tabular}
\caption{Held-out GCD diagnosis, every cell: T/F/N $=$ TP/FP/FN
identification, Ex $=$ exact master restorations, Rec.\ $=$
injected-overhead recovery on total power (\%).  The agent diagnostic is
perfect on both in all five seeds; blanket downsizing matches identification
but recovers only partially; random selection and stock repair introduce
false edits and negative recovery.}
\label{tab:gcd-grid}
\end{table*}

Three patterns are visible only at this resolution.  First, the blanket
arm's identification is perfect in every seed yet its restoration is
almost never exact: the gap between finding a defect and undoing it is
per-seed universal, not an aggregate artifact.  Second, the random arm's
recovery is negative in all five seeds: matched edit count without
diagnosis reliably makes the design worse.  Third, the agent's recovery beats the blanket arm's within every seed;
only across seeds does the blanket arm's best ($+31.9\%$, Seed 5) exceed
the agent's worst ($+29.1\%$, Seed 4), and on that same Seed 5 the agent
recovers $+51.1\%$.  The agent is also the only method with zero wrong
edits anywhere in the grid.

\section{Scope and Claim Boundaries}

With the raw evidence now on the table, this section draws the line
around what it supports.
The main paper claims mechanism evidence on a pinned engine rather than
statistical superiority; this section states the boundaries that claim rests
on.
The controlled $N{=}5$ campaign uses five seeds of one GCD foundation, not five
designs or public benchmarks.  It validates a frozen diagnostic and target depth,
not the causal benefit of a live LLM or delegation.  The first campaign execution
had an infrastructure handoff failure and required a pre-scored, sealed recovery.
The development visual sample is two tasks; UART is controlled and I2C has a
three-way tie; both visual-only edits fail DRC.  The three sealed held-out
campaigns gate only on exact placement (the M0 gate of the cross-design
table), five anchors each, and the contested margin is one
arm at $N{=}5$.  The natural I2C closure is adaptive development: full DRT
diagnosed the defect, but edits were replayed from a GRT checkpoint and fully
rerouted; the matched nonvisual and no-move controls of the cross-design
table (Table~\ref{tab:crossdesign}) were executed post hoc with
\ecoroute{}, so the accepted I2C
trajectory itself remains a development run.  The rank-2 candidate's
exact-placement (M0) check was omitted contrary to
the sealed rule, although the executed rank-1 evidence remains valid.  The
interactive Codex session lacks a sealed model transcript and exact sampling
metadata, and the UART no-edit control failed.  The tight-clock decision-flip
band is an STA-only re-judgment of frozen routed artifacts, not a full
tight-clock implementation flow.  \engine{} uses a merged-flat 3D encoding rather
than native z-coordinates.  Results are therefore mechanism evidence on a pinned
engine, not statistical superiority over OpenROAD or Pin-3D.

Every claim in the main paper is scoped to these boundaries by
construction; no generality claim is made or needed here.  Macro-free
RISC-V designs, public 3D benchmarks, complete tightened-constraint
flows, and sealed live-delegation ablations are the natural scale-out of
the same frozen contract.

\section{Extended Related Work}

The main paper's introduction places this work against agent, hardware-LLM,
and 3D-backend literature in one paragraph and points here for the full
survey.

\paragraph{General and iterative agents.}
General agents interleave reasoning with tools, learn API use, execute code,
and operate computer interfaces
\cite{react,toolformer,toollm,agentbench,codeact,sweagent,osworld}.  Iterative
methods use self-feedback, tool-grounded critique, search, or learned rules to
improve later behavior
\cite{selfrefine,reflexion,critic,lats,toolplanner,automanual}; specialized
agents can also calibrate one another's tool actions \cite{conagents}.  These
works do not address exact edits and externally verified acceptance in a
physical-design database.

\paragraph{LLMs and agents for hardware.}
LLMs for hardware first targeted conversational co-design, domain adaptation,
RTL generation and evaluation, accelerator generation, and tool-feedback
repair \cite{chipchat,chipnemo,llm4eda,verigen,rtllm,gpt4aigchip,
autochipfeedback}.  Agentic systems now coordinate multiple roles for RTL
generation and EDA scripts, interact with physical-design tools, optimize
flows, patch engine source, and support analog layout
\cite{chateda,edaid,mage,jarvis,openroadagent,layoutcopilot,orfsagent,
audopeda,openllmeco,pdagentbench,agenticedasurvey}, while vendors ship
multi-tool automation \cite{chipstack,agentengineer,siemensfuse}.
Post-route repair of design-rule violations is an established ECO task in
its own right, most recently for back-side routing in advanced
double-sided nodes \cite{ecobsdrv}.  Learned ECO
has used reinforcement learning in a fixed 2D action space \cite{irawareeco}.
Most closely, EvoDRC decomposes 2D block layouts into
bounded crops and evolves layer-wise repair skills from verified
operation--DRV records \cite{evodrc}.  It directly edits layout polygons and
uses local DRC plus a benchmark connectivity checker; it does not execute
routed 3D ECOs or enforce cross-tier preservation and hybrid-bond constraints.

\paragraph{Agentic EDA execution.}
LLM agents solve complex tasks more effectively when they reason, call tools,
inspect execution results, and revise decisions over multiple steps rather than
emit one static output.  In EDA, an academic line has begun to execute flows and
edit tools: ChatEDA connects an LLM planner to EDA tools \cite{chateda};
ORFS-Agent tunes flow parameters \cite{orfsagent}; AuDoPEDA uses coding agents to
improve OpenROAD source \cite{audopeda}; Retrieve--Schedule--Reflect studies
retrieval and optimization scheduling \cite{openllmeco}; and PDAGENT-BENCH
characterizes both language and vision agents for physical design
\cite{pdagentbench}.  OSWorld
highlights execution-grounded evaluation for multimodal agents in open-ended
computer environments \cite{osworld}.  In industry, Cadence introduced ChipStack
for AI-assisted chip design and verification \cite{chipstack}, Synopsys
introduced AgentEngineer for multi-agent automation of RTL and verification
tasks \cite{agentengineer}, and Siemens EDA launched the Fuse EDA AI Agent for
multi-tool automation from design to physical implementation \cite{siemensfuse}.
The open and 3D line provides the substrate: OpenROAD is the open physical
engine \cite{openroad}; Compact-2D and Pin-3D establish merged multi-tier
methodologies and 3D optimization \cite{compact2d,pin3d}; Open3DBench broadens
open 3D evaluation \cite{open3dbench}; and recent work uses LLMs for 3D-IC
space planning \cite{llm3dplan}.  These systems act at the level of flows,
parameters, code, or documents.  Surgical ECO on a true 3D backend, with
agent-selected observations and algorithms and with causal attribution of the
outcome to a specific component, has to our knowledge not been demonstrated.

\end{document}